\documentclass[compsoc, conference, a4paper, 10pt, times]{IEEEtran}

\usepackage{cite}
\usepackage{amsmath,amssymb,amsfonts}
\usepackage{algorithmic}
\usepackage{graphicx}
\usepackage{textcomp}
\usepackage{xcolor}
\usepackage{booktabs}
\usepackage[hidelinks]{hyperref}
\usepackage{multirow}
\usepackage{subcaption}
\usepackage{balance}

\makeatletter
\newcommand{\newlineauthors}{%
  \end{@IEEEauthorhalign}\hfill\mbox{}\par
  \mbox{}\hfill\begin{@IEEEauthorhalign}
}
\makeatother

\begin{document}

\title{ENCODE: Encoding NetFlows for Network Anomaly Detection}



\author{\IEEEauthorblockN{Clinton Cao}
\IEEEauthorblockA{\textit{Delft University of Technology} \\
Delft, Netherlands \\
c.s.cao@tudelft.nl}
\and
\IEEEauthorblockN{Agathe Blaise}
\IEEEauthorblockA{\textit{Thales SIX GTS France} \\
Gennevilliers, France \\
agathe.blaise@thalesgroup.com}
\and
\IEEEauthorblockN{Filippo Rebecchi}
\IEEEauthorblockA{\textit{Thales SIX GTS France} \\
Gennevilliers, France \\
filippo.rebecchi@thalesgroup.com}
\newlineauthors
\IEEEauthorblockN{Annibale Panichella}
\IEEEauthorblockA{\textit{Delft University of Technology} \\
Delft, Netherlands \\
a.panichella@tudelft.nl}
\and
\IEEEauthorblockN{Sicco Verwer}
\IEEEauthorblockA{\textit{Delft University of Technology} \\
Delft, Netherlands \\
s.e.verwer@tudelft.nl}
}


\maketitle

\begin{abstract}
    NetFlow data is a popular network log format utilized by many network analysts and researchers. Compared to deep packet inspection, NetFlow is easier to collect, easier to process, and less intrusive to privacy. Numerous studies have used machine learning (ML) to detect network attacks using NetFlow data. Typically, the initial step in these ML pipelines is to preprocess the data before passing it to the ML algorithm. Several approaches exist to preprocess NetFlow data; however, these often apply existing methods to the data, not considering the specific properties of network data. We argue that for data originating from software systems, such as NetFlow or software logs, similarities in frequency and contexts of feature values are more important than similarities in the value itself.

    In this work, we propose an encoding algorithm (ENCODE) that directly takes the frequency and context of the feature values into account when the data is being processed. Our algorithm clusters similar types of network behavior based on the computed frequencies and context of the feature values, aiding the process of detecting anomalies within the network. We empirically evaluate the effectiveness of our encoding on three public NetFlow datasets. We encode each dataset using our algorithm and train several ML models for network anomaly detection using the encoded data. Our evaluation reveals that ML models benefit from using our encoding for network anomaly detection.
\end{abstract}

\section{Introduction}\label{sec:introduction_encode}
NetFlow is a well-known format introduced by Cisco~\cite{CiscoSystems2018} in 1996, enabling network administrators to analyze the traffic occurring in the network. Each flow contains network statistics for a connection between two hosts. These statistics enable an administrator to infer whether any strange behaviors are occurring within the network. Using NetFlow data over deep-packet inspection for network analysis comes with multiple advantages; NetFlow is easier to collect, easier to process, and less intrusive to privacy, since it is not possible to inspect every packet in detail. However, NetFlow data comes with limitations. Because NetFlow only reports the summarized statistics of a communication between two hosts, crucial details may be lost, making it more challenging to infer malicious network. 

Several NetFlows datasets are publicly available, containing benign and malicious flows~\cite{Garcia2014_an_empirical,Macia-Fernandez2018_ugr,Moustafa2015_unsw,cic-ids}. The datasets can be used to train and evaluate the effectiveness of machine learning (ML) algorithms for the detection of network attacks. Before training a ML algorithm on a given dataset, the usual practice is to investigate and preprocess the dataset. This process helps remove any properties that could influence the final model learned from the data. Additionally, this process can help find the most important features for learning accurate models. Several techniques can be applied to preprocess a NetFlow dataset, including normalization of the numerical features~\cite{Larriva-Novo2020_efficient,Terzi2017_big_data,Yilmaz2019_expansion}, feature selection~\cite{Piskozub2019_malalert,Shamshirband2019_a_new_malware,Zoppi2021_meta_learning}, encoding numerical features~\cite{Camacho2019_semi_supervised,Magan-Carrion2020towards,Pellegrino2017_learning}, and extracting aggregated features from existing dataset~\cite{Ahmed2020_deep_learning,Haghighat2021_intrusion_detection,Nguyen2019_gee,Nugraha2020_performance}.

Most of these techniques treat the NetFlow data like a traditional data source, encoding the number of transferred bytes, number of transferred packets, and flow duration as standard numerical features. The similarity between flows is determined based on the similarity in the numerical values of these features. We believe that flows generated by different (software) services are not uniform and the similarity between flows should not be determined solely based on the raw feature values. For instance, similarly sized flows (e.g. based on the number of transferred bytes) can have entirely different meanings due to variations in the frequency distribution of their feature values. To address this, we propose a new NetFlow encoding technique (ENCODE) inspired by Word2Vec~\cite{Mikolov2013_efficient}, which treats every unique feature value as a different ``word". The similarities between these ``words" (i.e., unique feature values) are determined by their context and frequency. Specifically, we consider features values to be similar based on two criteria. First, the feature values must occur in similar frequencies. Second, the frequencies of the previous and subsequent values (extracted from the previous and subsequent flows) must have a similar distribution. Similar to Word2Vec, our encoding algorithm constructs a co-occurrence matrix to store all the computed frequencies. Each row of this matrix represents the context computed for a unique feature value. We cluster the rows to find groups of values exhibiting similar behavior and use the cluster labels to encode values of the corresponding feature. Our encoding algorithm computes a co-occurence matrix and a set of cluster labels for each input NetFlow feature.  Once the given input features are encoded, the resulting encoded dataset is constructed and passed to the ML algorithm for training. 

To evaluate the effectiveness of our algorithm, we preprocessed NetFlow data from three different datasets. Our encoding was applied to three flow features: duration, number of transferred bytes, and number of transferred packets. The protocol was included as a discrete feature, while all other features, such as the used ports, were ignored for training. Although ports can be useful for detecting network intrusions, they frequently contain spurious patterns due to the data collection setup. Our experiments include a new dataset created for detecting anomalies in a Kubernetes cluster. Across all datasets, our encoding consistently improved the performance of several ML models in the network anomaly detection tasks, suggesting that ML models benefit from our encoding. 

Our evaluation inidicates that state machine models experienced the greatest performance improvment when using our encoding. For this method, we provide a practical use-case. We describe what an analyst receives as output from the system and how they should use it to infer an attack is taking place. While a sophisticated attacker might attempt to evade detection, this is not straightforward. An attacker who randomizes or systematically modifies the generated flows is still detected by our system. In fact, such modifications often make attack flows more anomalous, thereby simplifying their detection. The attack could avoid detection by mimicking the benign flows used during training, but this requires an undetected pre-existing attack. Moreover, truly mimicking benign behavior would bypass any ML-based detector.

\subsection{Our Contribution}
Our key contributions in this work are summed up as follows:
\begin{itemize}
    \item We provide a new encoding algorithm that can be used to preprocess NetFlow data. The algorithm extracts context and frequency information, and uses these to build clusters of feature values exhibiting similar behavior.
    \item We show an application of our encoding algorithm by training ML models from the encoded NetFlow data in an unsupervised manner and using it for anomaly detection.
    \item We provide a new NetFlow dataset that researchers can use to evaluate anomaly detection methods in a Kubernetes cluster. We also provide the testbed that researchers can use to create their dataset.
    \item We demonstrate that ML models benefit from using our encoding.
    \item We show how an anomaly detection system based on our encoding can be used in practice and is resilient to malicious data permutations.
\end{itemize}

\subsection{Outline of Paper}
The rest of this paper is structured as follows: we first describe the threat model that we have considered in this work in Section~\ref{sec:threat_assumptions_encode}. Then we list some related works in Section~\ref{sec:related_works_encode}. We provide the intuition behind our encoding in Section~\ref{sec:intuition_encode}. In Section~\ref{sec:encoding_algorithm_encode}, we describe our NetFlow encoding algorithm. Then in Section~\ref{sec:evaluation_encode}, we describe our procedure in how we evaluate our new algorithm, and the empirical results of our evaluation are presented in Section~\ref{sec:empirical_results_encode}. In section~\ref{sec:robustness_encode}, we discuss a use-case and the robustness of our encoding. Finally, we conclude this paper and provide future work in Section~\ref{sec:conclusion_future_encode}.
\section{Threat Model}\label{sec:threat_assumptions_encode}
In this work, we consider a specific threat model for network anomaly detection using models learned from ML algorithms. We briefly describe the goals and the capabilities of the adversary.

\subsection{Goals of the Adversary}
In our threat model, we consider an adversary aiming to launch network attacks against a target network for various objectives, such as infiltrating the network to find sensitive data, taking down a network through a denial-of-service (DoS) attack, operating a botnet, and installing malware. Furthermore, the adversary seeks to evade any attack detection system deployed within the network. However, the adversary does not have access to the network or the detection system's rules and models.

\subsection{Capabilities of the Adversary}
We assume that the adversary is aware of the presence of a ML-based anomaly detection system within the targeted network. However, they lack information about the specific models or algorithms used within the anomaly detection system. The adversary does not have the capabilities to deploy probes within the network to monitor the flows occurring in the network. Moreover,  adversary can attempt to evade detection by manipulating the flows they generate, but cannot make these flows appear benign through direct analysis of the anomaly detection system of by studying the benign network data.
\section{Related Works}\label{sec:related_works_encode}
There are various techniques for preprocessing NetFlow data to train ML algorithms. One common technique involves normalizing the numerical (non-categorical) feature data using the Z-score method~\cite{Larriva-Novo2020_efficient,Terzi2017_big_data,Yilmaz2019_expansion}. This method scales numerical feature values to fall within the range of 0 to 1, removing large differences in the feature data that could have a large impact on model prediction results. Normalization ensures that each (numerical) feature would have an equal impact on the model prediction results. However, one drawback of this method is that behavioral patterns present in the NetFlow data may be lost. The normalization is performed by first computing the mean of the feature values for a given input feature, which assumes that there is an average behavior shared between all hosts in the network. This assumption may not hold in networks with diverse software services, as each may exhibit distinct behaviors tied to specific contexts. Consequently, this method may remove important contextual information, potentially reducing the model's accuracy.  

Another technique to preprocess NetFlow data is to rank and select features using various methods~\cite{Piskozub2019_malalert,Shamshirband2019_a_new_malware,Zoppi2021_unsupervised}. These methods focus on using features that provide a substantial contribution to the model prediction results, reducing the number of features needed to train an accurate model. However, these methods do not explicitly use the frequencies and context of the feature values to create an encoding for each flow.

Besides normalization and feature selection, there are techniques that create an encoding for numerical feature values~\cite{Camacho2019_semi_supervised,Magan-Carrion2020towards,Pellegrino2017_learning}. These methods divide the range of a numerical feature into discrete bins and encode each feature value based on based on its corresponding bin. However, these encoding techniques ignore the frequencies of the features values and the context in which they occur.

Several studies compute aggregated features from the existing NetFlow features and use these to train ML algorithms~\cite{Ahmed2020_deep_learning,Haghighat2021_intrusion_detection,Nguyen2019_gee,Nugraha2020_performance}. These aggregates can reveal new information from the dataset,  potentially enhanching the training of a ML model. In our work, we also compute new aggregated features to capture the context in which specific feature values occur and use solely these new features to train a model. Compare to the previous studies~\cite{Ahmed2020_deep_learning,Haghighat2021_intrusion_detection,Nguyen2019_gee,Nugraha2020_performance}, we do compute fewer aggregated features and ensure there is no (semantic) overlap among these features. Additionally,  these studies do not explicitly use frequencies of the feature values as an aggregated feature, nor do they apply an encoding to the (numerical) feature data. 

\section{Intuition Behind ENCODE}\label{sec:intuition_encode}
We believe the context and frequency of a unique feature value provide valuable information about the type of network behavior occurring in a flow. To illustrate this, consider the following example: suppose we have extracted the frequencies of the following byte values from NetFlow data: 37, 39, 80, 81, and 37548. The frequencies of the byte values are presented in Table~\ref{tab:bytes_frequency_example}. Based on the extracted frequencies, byte values 80 and 81 differ only by one byte in size, yet their frequencies vary substantially. This substantial difference likely indicates that they are generated by different type of software services and should therefore be treated differently by ML methods.

Using a NetFlow preprocessing technique such as the one employed by Camacho et al.~\cite{Camacho2019_semi_supervised}, the range of the byte values can be divided into three regions; 37 and 39, 80 and 81, and 37548. Despite the substantial difference in the frequencies of byte values 80 and 81, they are placed within the same region. This causes ML algorithms to treat these two values as similar. 

In contrast, ENCODE divides the range of byte values based on their corresponding frequencies. Byte values with a very low frequency are placed together into one region, while byte values 80 and 81 are each assigned to their own separate region. This results into three different regions: the first region contains bytes values 37, 39, and 37548. The second region contains solely byte value 80, and third region contains solely byte value 81. With this separation, ML algorithms can now treat bytes values 80 and 81 as two distinct values.

\begin{table}[h]
\centering
\caption{Frequencies of byte values 37, 39, 43, 80, and 81. Frequencies are extracted from a subset of flows in the UGR-16 dataset.}\label{tab:bytes_frequency_example}
\begin{tabular}{@{}c|c@{}}
	\toprule
	Byte Value & Frequency \\ \midrule
	37         & 1         \\
	39         & 4         \\
	80         & 24771     \\
	81         & 3158      \\
	37548      & 4         \\ \bottomrule
\end{tabular}
\end{table}
\section{Encoding Algorithm}\label{sec:encoding_algorithm_encode}
In this section, we describe our encoding algorithm (ENCODE) for NetFlow data. We begin by explaining how the context is computed for each unique flow feature value and how this is used to create its encoding. Then we provide the implementation details of our encoding algorithm. ENCODE is publicly available at our GitHub repository~\cite{encode_repo}. ENCODE takes as input a set of numerical (non-categorical) features with their corresponding values. Its output is a mapping between each feature value and its corresponding encoded feature value.

\subsection{Computing the Context}
Our method for computing the context of values from the provided input feature is inspired by the work of Word2Vec, particularly the Count Bag of Words (CBOW) method~\cite{Mikolov2013_efficient}. For each value $v$ of an input (numerical) NetFlow feature $f$, we compute the context in which $v$ occurs by computing the frequencies of the values that occurred immediately before and after $v$. The intuition is that these frequencies capture the behavior of $v$ within the network i.e. which other feature values $v$ co-occurs with and how often.

The computed frequencies are used to generate a vector for value $v$. Before computing the context frequencies for each value $v$, the flows are grouped by connection (i.e. grouped by the pair of source and destination hosts). This ensures the context of $v$ is not influenced by other flows co-occurring in the same network. As a result, the context represents the surrounding flows generated by similar software services.

\subsection{Creating the Encoding}
From the context generation, we obtain vectors representing the co-occurrence frequencies of feature values. We use the Euclidean distance to determine whether two feature values $v$ and $v'$ occur in similar contexts. A smaller distance between two vectors indicates a higher similarity in the contexts of the two corresponding feature values. This enables us to create groups of feature values that are contextually similar.

The encoding for the feature values is created by clustering the vectors of these values, with the assigned cluster labels used as their encoding.  Thus, the encoding for $v$ is the corresponding cluster label that was assigned to its vector. Feature values with similar frequencies and contexts are assigned to the same cluster and receive the same encoding. For the clustering of the vectors, we employ the KMeans algorithm because it allows us to control the number of clusters that can be formed, thereby also giving us control over the size of the encoding for each input feature. While we have opted to use KMeans as our clustering algorithm, other clustering algorithms can also be used to compute the encoding. ENCODE is modular by design, allowing the clustering algorithm to be easily replaced with one that better fits the user's needs.

\subsection{An Illustrative Example}
\begin{figure}
    \centering
    \includegraphics[scale=0.5]{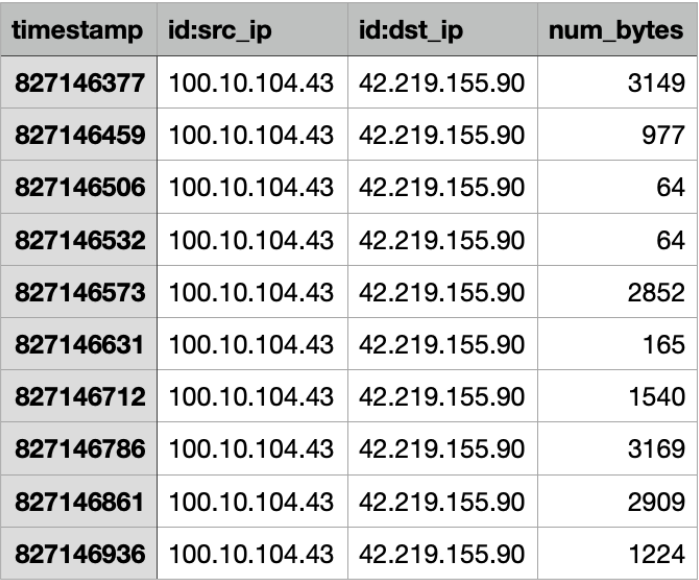}
    \caption{A small subset of NetFlow data extracted from the UGR-16 dataset. }
    \label{fig:ugr_small_netflow_sample}
\end{figure}

We now provide a small illustrative example to demonstrate how our encoding is computed for NetFlow data. The flows used in this example are a very small subset of flows extracted from the UGR-16 dataset, one of the datasets used in our evaluation. Figure~\ref{fig:ugr_small_netflow_sample} presents the small subset of flows used in this example. More information on this dataset can be found in Section~\ref{sec:evaluation_encode}. In this example, we demonstrate the computation of the context and the encoding for unique byte values. 

\begin{figure}[t]
    \centering
    \includegraphics[width=\linewidth]{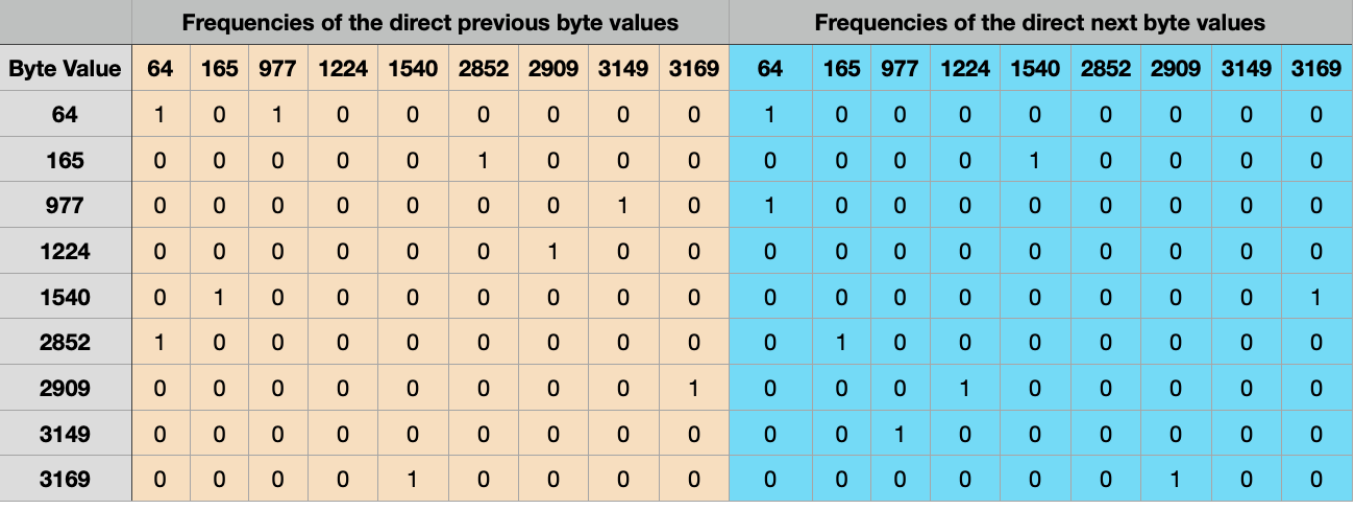}
    \caption{ Frequencies of the direct previous and next bytes values for each of the unique byte values that are present in Figure~\ref{fig:ugr_small_netflow_sample}.}
    \label{fig:ugr_small_example_frequency}
\end{figure}

\begin{figure}[t]
    \centering
    \includegraphics[scale=0.5]{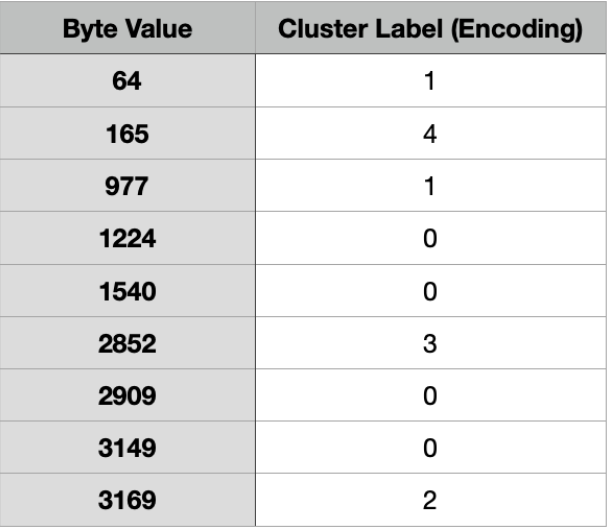}
    \caption{Cluster labels assigned to the unique byte values used in the example. The assigned cluster labels are used as the encoding for the unique byte values. In this example, we have used five clusters to cluster the vectors.}
    \label{fig:ugr_small_example_encoding}
\end{figure}

The first step of ENCODE is to find the unique byte values. From the byte values that are shown in Figure~\ref{fig:ugr_small_netflow_sample}, we observe a total of nine unique byte values: 64, 165, 977, 1224, 1540, 2852, 2909, 3149, and 3169. Note that all data in this example comes from the same connection. Next, we compute the frequencies of the direct previous and next byte values for each of these nine byte values. The computed frequencies are shown in Figure~\ref{fig:ugr_small_example_frequency}. Each row in Figure~\ref{fig:ugr_small_example_frequency} represents the vector compued for the corresponding unique byte value. These vectors are then clustered together using the KMeans algorithm and the assigned cluster labels serve as the encoding for the corresponding unique byte values. Figure~\ref{fig:ugr_small_example_encoding} presents the assigned cluster labels for each unique byte value.

\subsection{Implementation Details}
\begin{figure}
    \centering
    \includegraphics[width=\linewidth]{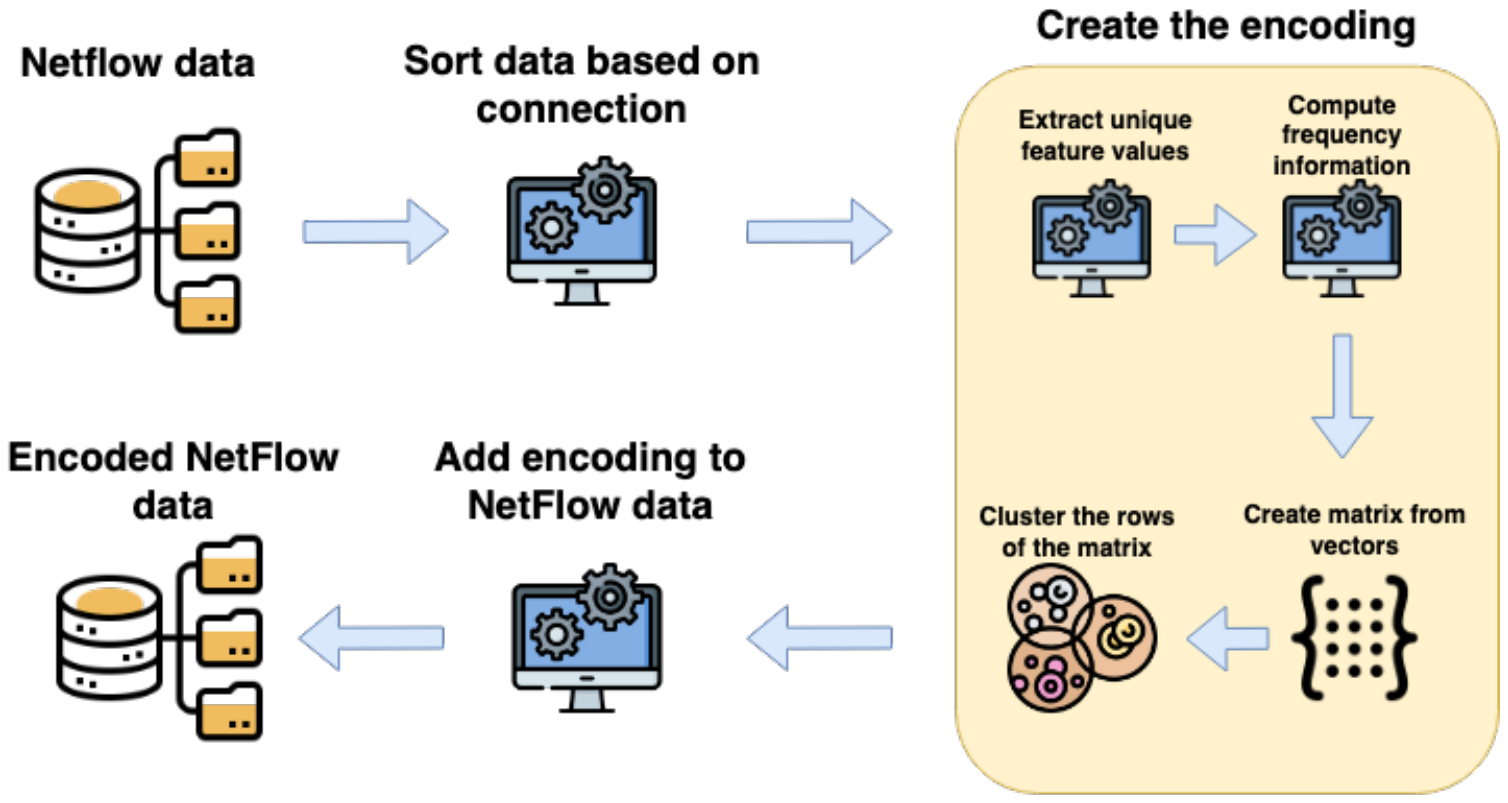}
    \caption{High-level overview of our data processing pipeline. The pipeline shows how the encoding is created for the given NetFlow input data using our encoding algorithm.}
    \label{fig:data_processing_pipeline}
\end{figure}

Figure~\ref{fig:data_processing_pipeline} provides a high-level overview of ENCODE's data processing pipeline. In the following pararaphs, we describe in detail the challenges we faced when constructing the co-occurrence matrix within ENCODE and how we addressed the challenges. Furthermore, we also describe how we cluster the rows of the co-occurrence matrix and how we evaluate the quality of the formed clusters for the encoding. 

\subsubsection*{\textit{Generalized Vector Structure}}
As shown in our example, a vector is generated for each unique feature value, $v$. Since these vectors are clustered using KMeans, the sizes of the vectors must be uniform. Additionally, all vectors must store the same type of information (co-occurrence frequency values corresponding to a particular feature value) in the exact same positions. 

As illustrated in Figure~\ref{fig:ugr_small_example_frequency}, the first half of each vector (highlighted in orange) represents the frequencies of feature values occurring directly before $v$, while the second half (highlighted in blue) represents the frequencies of feature values occurring directly after $v$. The size of each vector is equal to twice the number of unique feature values, ensuring that all vectors share the same structure and size. However, this structure comes with several limitations: 

\begin{itemize}
    \item It can lead to the creation of many sparse vectors. 
    \item The size of vectors depends on the number of unique values for a given feature. As the number of unique values increases substantially, the size of the vectors also expands proportionally.
    \item We cannot capture the behavior in which a feature value $v$ co-occurs frequently with itself. It is common to observe $v$ co-occur with itself in NetFlow data, indicating repetitive software behavior.
\end{itemize} 

The last limitation is somewhat nuanced. The co-occurences of a value $v$ with itself are present in the table. For instance, for value 64 in Figure~\ref{fig:ugr_small_example_frequency}, the co-occurences with itself are stored in the column labeled 64. The challenge is that,  for different data rows, these self-contexts (co-occurence with itself) are located in different columns. Consequently, rows with only self-contexts would have a large distance from rows showing a similar pattern. In our view, such repetitive behavior reflects similar software activity (behavior) and similarity should not depend on the exact feature values used in the flow. 

To address the last limitation, we add two separate columns to store the ``self-frequency" i.e., one column for $v$ occurring directly before itself and another for $v$ occurring directly after itself. In addition to storing the self-frequency, we have also included an additional entry that stores the logarithm of the frequency of $v$ (marked in red in Figure~\ref{fig:ugr_small_example_frequency}). This way the frequency of $v$ also influences the distance computation between vectors. Furthermore, this entry helps to differentiate features that have similar previous (orange) and next (blue) values but different frequencies.

\begin{figure}[t]
    \centering
    \includegraphics[width=\linewidth]{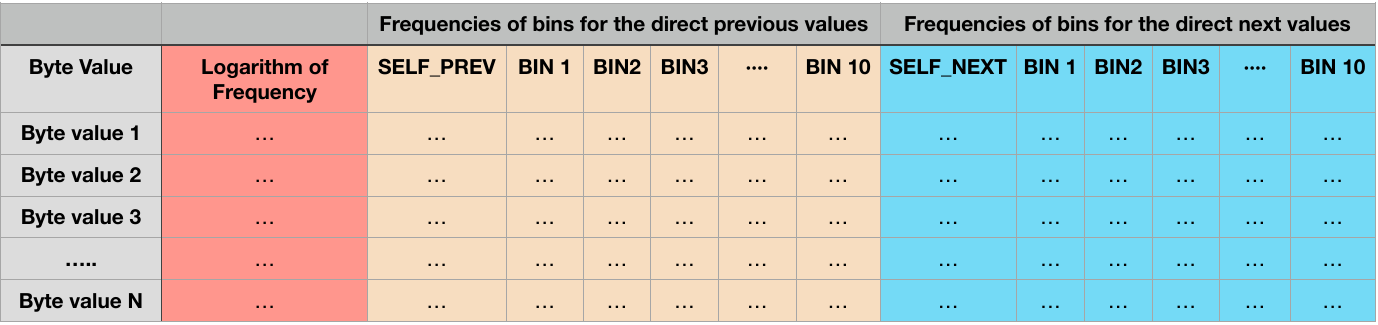}
    \caption{Generalized vector structure used for each unique feature value within ENCODE.}
    \label{fig:encoding_final_vector_structure}
\end{figure}

To resolve the first two limitations, we divide the space of the previous and next values into ten bins each using percentiles. This results in a vector of length 20 for each feature value $v$. A benefit of this solution is that it enables us to control the size of the vectors by adjusting the number of bins used to divide the space of the previous/next values. Figure~\ref{fig:encoding_final_vector_structure} presents the final generalized structure used in the implementation of ENCODE.

\subsubsection*{\textit{Cluster Evaluation within ENCODE}}
To cluster the rows of the co-occurence matrix constructed by ENCODE, we use the KMeans implementation provided by the scikit-learn Python package~\cite{sklearn_kmeans}. We opted for KMeans as it allows us to control the number of clusters, thereby fixing the size of our encoding.  Since KMeans selects the initial centroids at random, each run could lead to different final clusters. ENCODE therefore runs KMeans ten times and selects the best run to create the encoding for the input feature. The Silhouette Coefficient is used to evaluate the quality of the formed clusters (of the vectors). To compute the Silhouette Coefficient, we make use of scikit-learn's Python package~\cite{sklearn-silhouette}.

\section{Evaluation of ENCODE}\label{sec:evaluation_encode}
In this section, we describe how we evaluate our encoding in the context of network anomaly detection. The goal of our evaluation is to show that ML algorithms leverage our encoding to extract contextual information from NetFlow data, enabling them to learn more accurate models for network anomaly detection. To this end, 
we train several ML models in an unsupervised manner to detect various network attacks. For each selected ML algorithm, we train two models; one that includes our encoding as part of the input features and one that does not. The performance of these two models is then compared. Specifically, we compare their performance using Balanced Accuracy and the $F_1$ score 

\subsection{Selection of ML Models}
To evaluate the effectiveness of ENCODE on learning ML models for network anomaly detection, we selected several ML methods for unsupervised network anomaly detection.

\subsubsection*{State Machine}
A state machine, formerly known as a finite state automaton, is a mathematical model used to represent the sequential behavior of a system. These models can be learned using a passive-learning approach that leverages state-merging algorithms. Since state machines inherently capture sequential behavior, they are well-suited to modelling NetFlow. Prior works have demonstrated the effectiveness of using state machines for network anomaly detection~\cite{Pellegrino2017_learning,Matousek_2021_Efficient,Grov_2019_Towards}. In the context of unsupervised network anomaly detection, we generate fixed-size behavioral sequences exclusively from benign flows to learn a state machine. The resulting model is used to detect any sequential behavior at test time deviating from the learned behavior. To learn a state machine from NetFlows, we use Flexfringe~\cite{Verwer2017_Flexfringe}, a framework that supports various merging heuristics for the learning of state machines.

\subsubsection*{Isolation Forest}
Isolation Forest (IF) is a well-known unsupervised anomaly detection method. The core idea is anomalies are easier to isolate because these tend to be farther from other data points within the feature space, resulting in shorter path lengths to the root of the tree.  Consequently,  instances with shorter average path lengths to the root nodes of the trees are considered more anomalous~\cite{Liu08_isolation}. In contrast to state machines, IF does not model sequential behavior and is specifically designed to detect point anomalies. For our evaluation, we use the implementation of IF provided by the scikit-learn Python package~\cite{sklearn-if}.

\subsubsection*{Local Outlier Factor}
Local Outlier Factor (LOF) is another well-known anomaly detection method. The core idea behind LOF is that instances grouped within a dense neighbourhood are considered to be benign, while those in sparsely populated neighborhoods are deemed. anomalous. For our evaluation, we use the implementation of LOF provided by the scikit-learn Python package~\cite{sklearn-lof}.

\subsubsection*{Deep Neural Network}
In recent years, deep neural networks have gained significant popularity for anomaly detection. DeepLog is a well-known work that uses an LSTM to detect anomalies from logs collected from a system~\cite{du2017deeplog}. Much like state machines, , LSTMs are used to model sequential behaviors that occur within a system. Furthermore, only benign data is used to train the model, which is then employed to detect sequential behaviors deviating from those observed during training. For our evaluation, we use the PyTorch implementation of DeepLog provided by van Ede et al.~\cite{deeplog_implementation_repo}.

\subsection{Comparison Between Encoding Methods}
To evaluate the effectiveness of ENCODE, we have selected two encoding methods to serve as baselines: an existing percentile-based encoding and a naive frequency-based encoding. The percentile-based encoding divides the range of a given input feature into $b$ bins, and the encoding of a feature $v$ corresponds to the bin that it falls into. The frequency-based encoding creates an encoding based on the frequency of the feature values. If a feature value $v$ occurs at least $x$ times, then the raw value is used as its encoding. Otherwise,  it is grouped into a low-frequency bin along with other feature values that do not meet the threshold of $x$. These low-frequency bins are further subdivided into multiple bins using the percentile-based encoding method.
 
\subsection{Dataset Selection}
\begin{figure}[!ht]
   \centering
   \includegraphics[width=\linewidth]{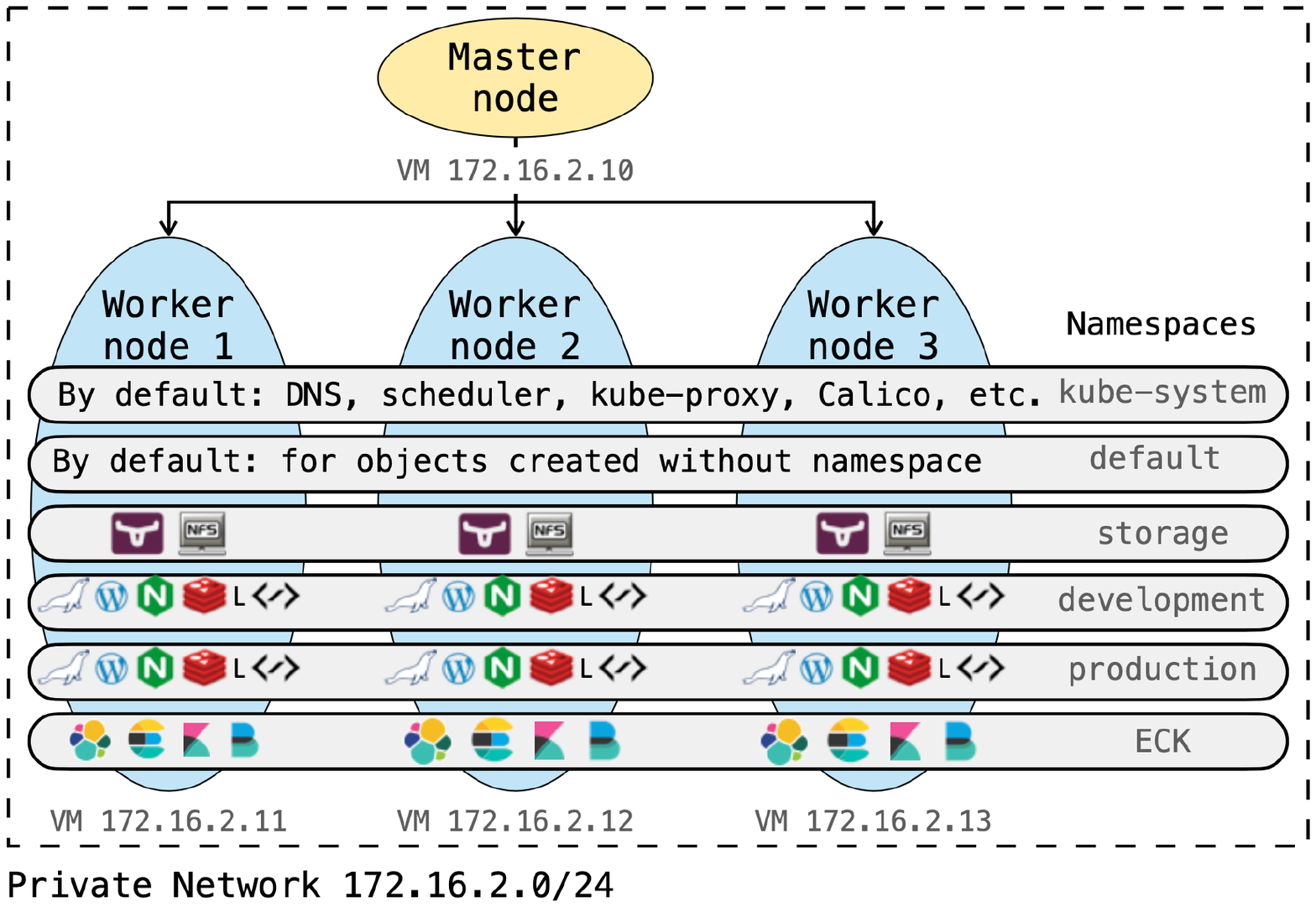}
   \caption{Architecture of the Kubernetes Cluster used to generate the AssureMOSS dataset. Image is taken from the work done by Cao et al.~\cite{Cao_2022_Learning}.}
   \label{fig:kubernetes_cluster}
\end{figure}

To assess the generalizability of ENCODE across different network architectures, we selected three datasets to train models for anomaly detection. The first dataset, the AssureMOSS dataset, is a new dataset created for the AssureMOSS project~\cite{assureMOSS_project}. It contains both benign and malicious NetFlows collected from a Kubernetes cluster running multiple microservice applications~\cite{assuremoss_dataset}.  The architecture of the cluster is shown in Figure~\ref{fig:kubernetes_cluster}. To generate benign flows for this dataset, 30 participants were recruited to use the microservice applications deployed within the cluster~\cite{Cao_2022_Learning}. Furthermore, three different attack scenarios were constructed (DoS attack, lateral movement, and malicious code deployment) using a threat matrix proposed by Microsoft~\cite{microsoft_threat_matrix}. These attack scenarios were launched against the cluster, while the participants were actively using the applications. 

As the second dataset, we selected a well-known dataset widely used in previous studies: the CTU-13 dataset, created by Garcia et al.~\cite{Garcia2014_an_empirical}. This dataset contains both benign and malicious flows. Benign flows were generated by real users on a university network, while the malicious flows were generated by various botnet malware. In total, the dataset comprises thirteen captures collected from different botnet samples.

For the third and final dataset, we selected the UGR-16 dataset created by Maciá-Fernández et al.~\cite{Macia-Fernandez2018_ugr}. We chose this dataset because it contains a substantial collection of NetFlows data, totaling to 100 GB. The flows were collected over a period of 6 months from a Spanish ISP.  The benign flows originated from real users accessing the internet for various purposes, while the malicious flows were produced by various types of network attacks.  

\subsection{Feature Selection}
We run ENCODE on the same feature set across all three datasets: the protocol used in a flow, the total number of bytes sent in a flow, the total number of packets sent in a flow, and the duration of the flow. Using only these four features, ENCODE does not rely on operating system- or network-specific features, making it applicable to various types of networks. We exclude the source and destination ports as a feature for ENCODE as these tend to be operating system or network-specific. Additionally, port information can sometimes act as a spurious feature, i.e., a feature that correlates with the class label without being truly indicative of the class.

As an example, we use the CTU-13 dataset to illustrate how the source port is a spurious feature of the dataset. We believe that the source port in the CTU-13 dataset leaks information about the flow's label and should not be blindly used to train a model. Garcia et al. generated the malicious flows using virtual machines running Windows XP. In this operating system, the default ephemeral ports range from 1025 to 5000~\cite{microsoft_xp_ports}.

\begin{figure}[!ht]
  \centering
  \subfloat[Source port distribution of benign flows]{
    \includegraphics[width=0.45\textwidth]{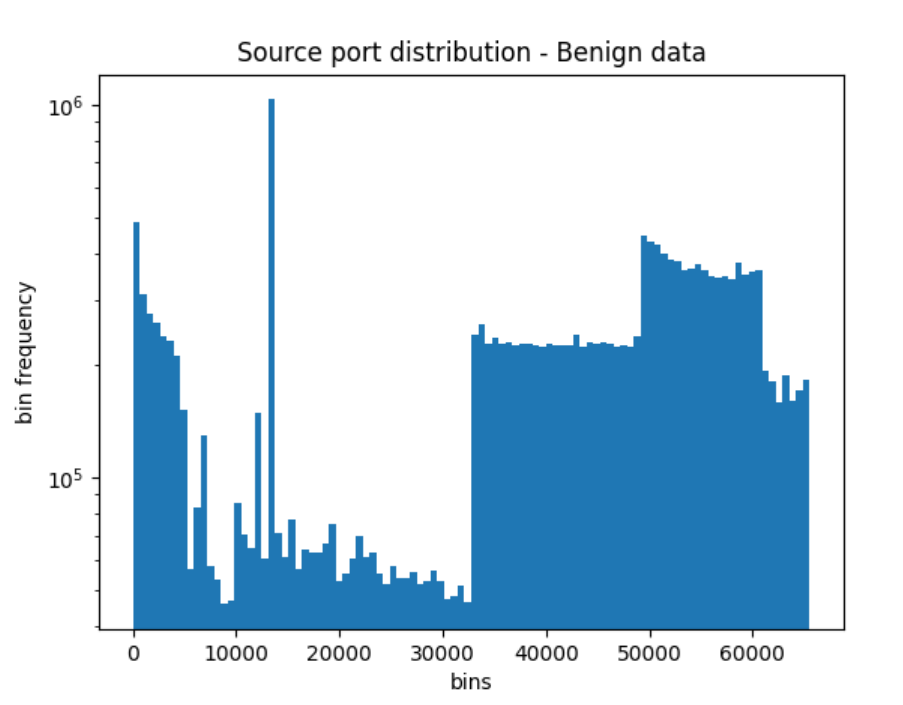}
  }
  \\
  \subfloat[Source port distribution of malicious flows]{
    \includegraphics[width=0.45\textwidth]{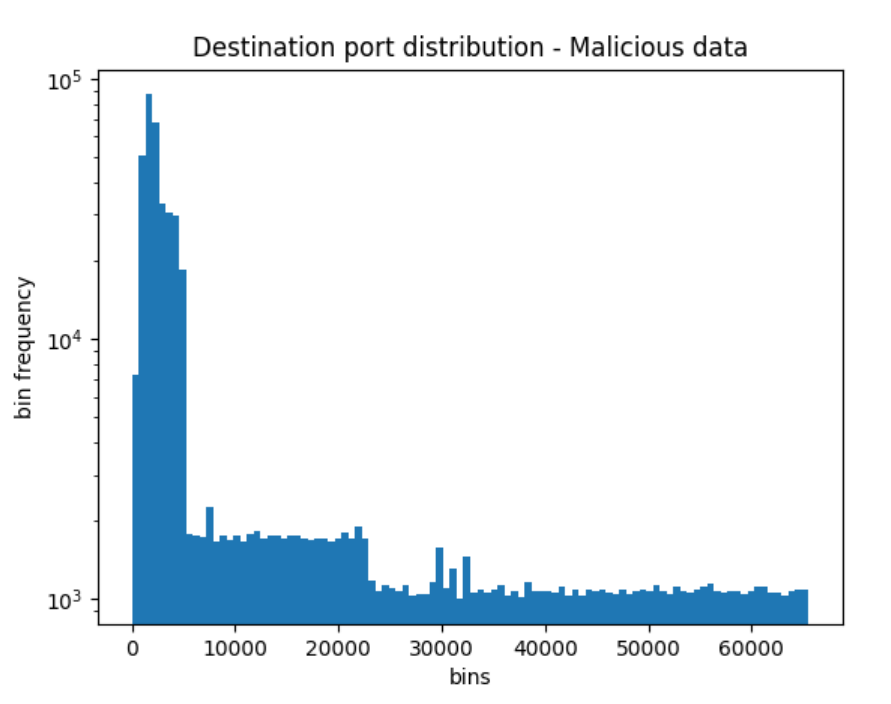}
   }
  
  \caption{Source port distribution in the CTU-13 dataset.}\label{fig:ctu_src_port_dist}
\end{figure}

By plotting the distribution of the protocols for both benign and malicious flows, we observe a clear difference in the distributions of the source ports. Figure~\ref{fig:ctu_src_port_dist} presents the source ports distribution for benign and malicious flows across all thirteen scenarios. Within the malicious flows, we observe a substantial portion of the flows appear to use a source port lower than 6000. The top three ports used in malicious flows are 2077, 1025, and 2079. All three source port numbers fall within the range used by the virtual machines. We believe that by training a model using the source port causes it to identify whether a flow originates from a virtual machine rather than determining if it is malicious or benign. Consequently, using the source port to train a model would falsely lead to a better model performance. Therefore, we opted not to run ENCODE on the source and destination ports of the flows.

\subsection{Encoding Train \& Test Data}
As we perform anomaly detection in an unsupervised manner, we use only benign flows as training data for the ML models. The test data can contain both benign and malicious flows. Below, we outline how we selected the training and test sets for each dataset.

For the CTU-13 dataset, we used the split proposed by the original authors of the dataset to create our training and test sets~\cite{Garcia2014_an_empirical}.  To construct the training set, we randomly sampled benign flows across all training scenarios, resulting in approximately 720000 benign flows. Similarly, we randomly sampled benign and malicious flows across the test scenarios, resulting in a test set containing approximately 522000 thousand flows. 

As for the UGR-16 dataset,Maciá-Fernandez et al. have provided the split between training and test data. We followed the same procedure as with CTU-13 to construct our training set. Specifically, we randomly sampled benign flows from the listed calibration data, resulting in approximately 600000 flows.  As for our test set, we randomly sample benign and malicious data from the listed test data, resulting in a test set containing approximately 664000 flows.

For the AssureMOSS dataset, Cao et al. provided the split between training and test data. The training data contains solely benign flows and has approximately 120000 flows.  The test data contains both benign and malicious flows and has approximately 275000 flows.  We directly use the training data as our training set and the test data as our test set. 

After creating the training and testing sets for each dataset, the training set, along with the selected features, is provided as input to ENCODE to learn an encoding for the corresponding dataset. The resulting encoding is then applied to the input features of the corresponding dataset and the encoded features are added as new columns to the training and testing sets.

\subsection{Creating Sequences For Sequential Models}
Since we are using two ML models that learn the sequential behavior from sequences, we need to transform the NetFlow data into sequences (or traces). Each sequence consists of a list of symbols, where each flow is transformed into a symbol. A symbol for an arbitrary flow is constructed by concatenating the encoded values of the selected features into a single string in the following format: $PROTOCOL\_BYTES\_PACKETS\_DURATION$. Note that the protocol feature is already a categorical feature and does not need to be encoded using ENCODE. Instead, we apply an integer encoding that converts each protocol into a numerical value. To generate traces, we use a sliding window of length 20 on the symbols constructed from flows of the same connection (i.e. combination of source and destination host). Traces shorter than the sliding window size are discarded, as we deem these insufficient to provide meaningful information for model learning.

\section{Results}\label{sec:empirical_results_encode}
\begin{figure*}[h]
    \centering
    \includegraphics[width=0.23\linewidth]{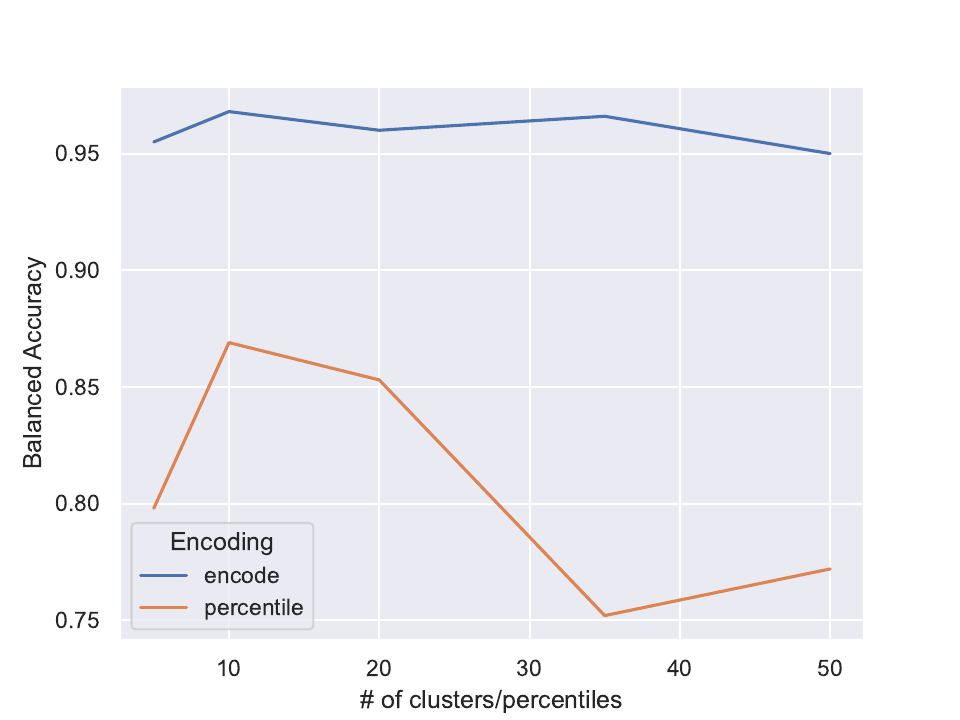}
    \includegraphics[width=0.23\linewidth]{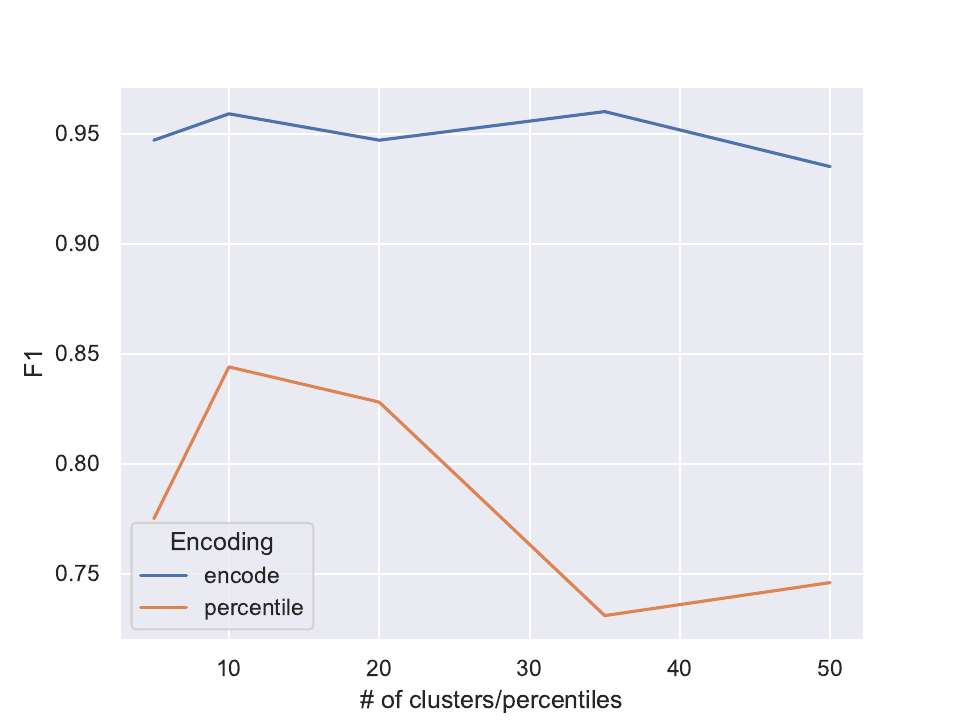}
    \includegraphics[width=0.23\linewidth]{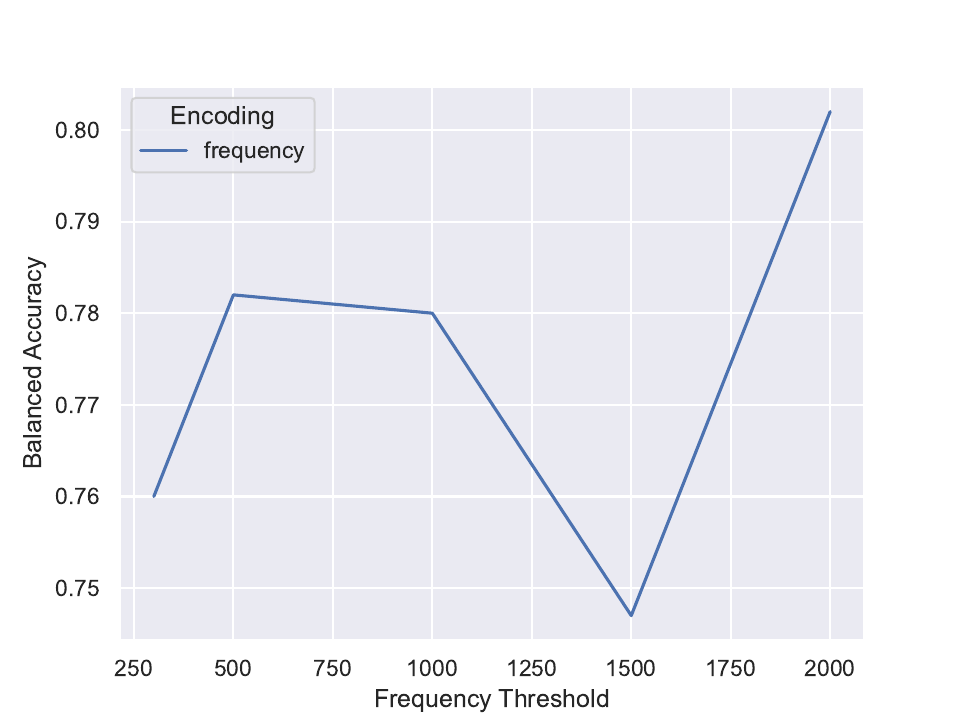}
    \includegraphics[width=0.23\linewidth]{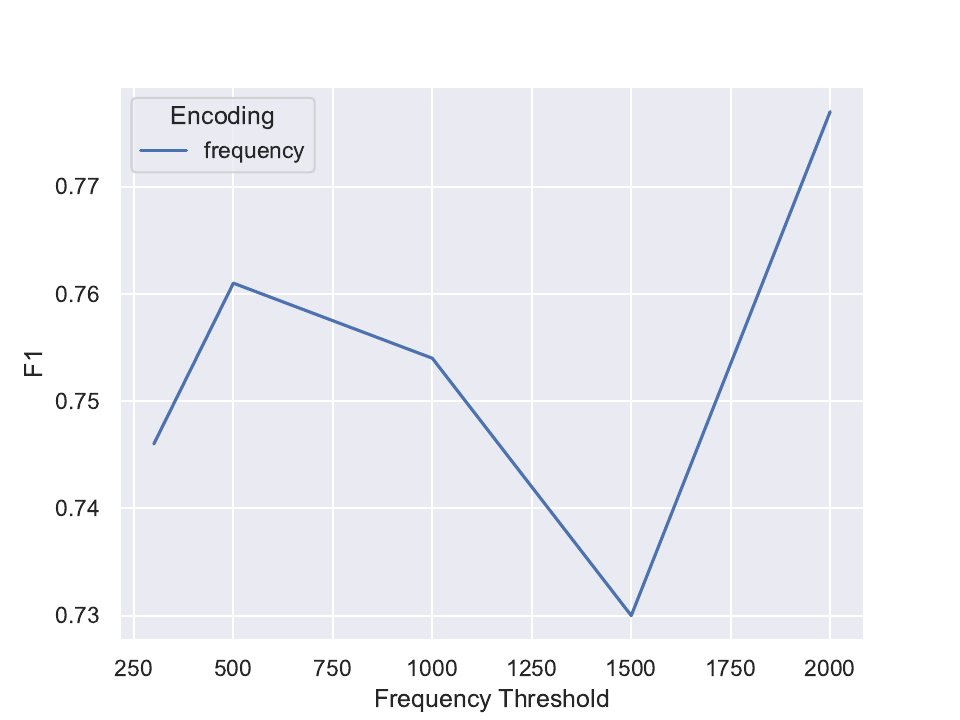}
        \caption{Performance results of State Machine (for each encoding method) in detecting anomalies on the AssureMOSS dataset. The figure presents the model’s average results over 10 runs.}
    \label{fig:experiment_results_assuremoss_sm}
\end{figure*}

\begin{figure*}[h]
    \centering
    \includegraphics[width=0.23\linewidth]{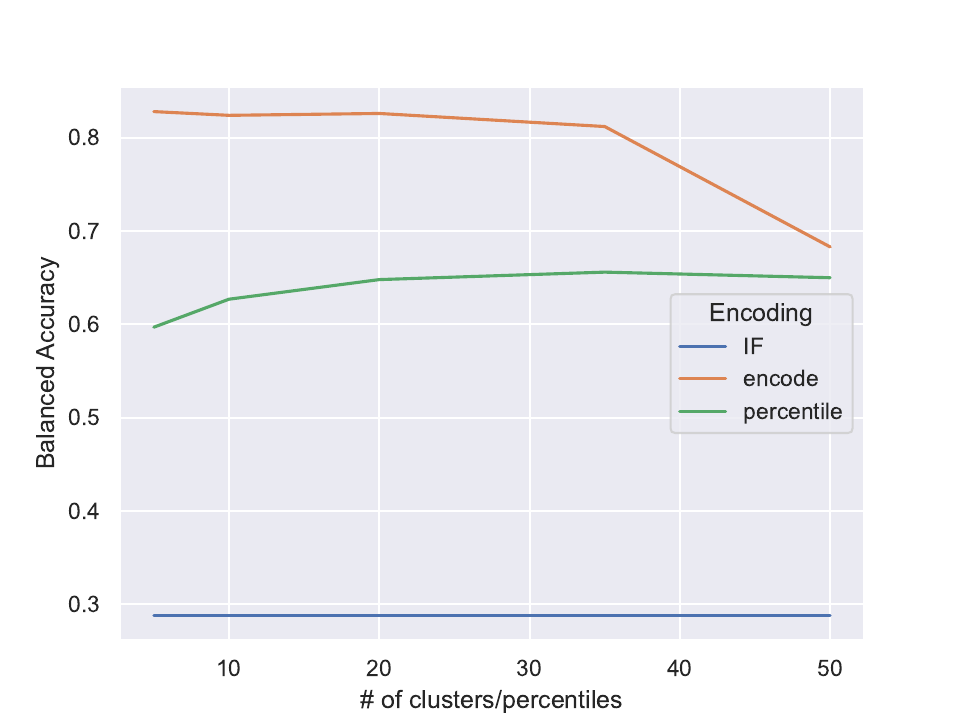}
    \includegraphics[width=0.23\linewidth]{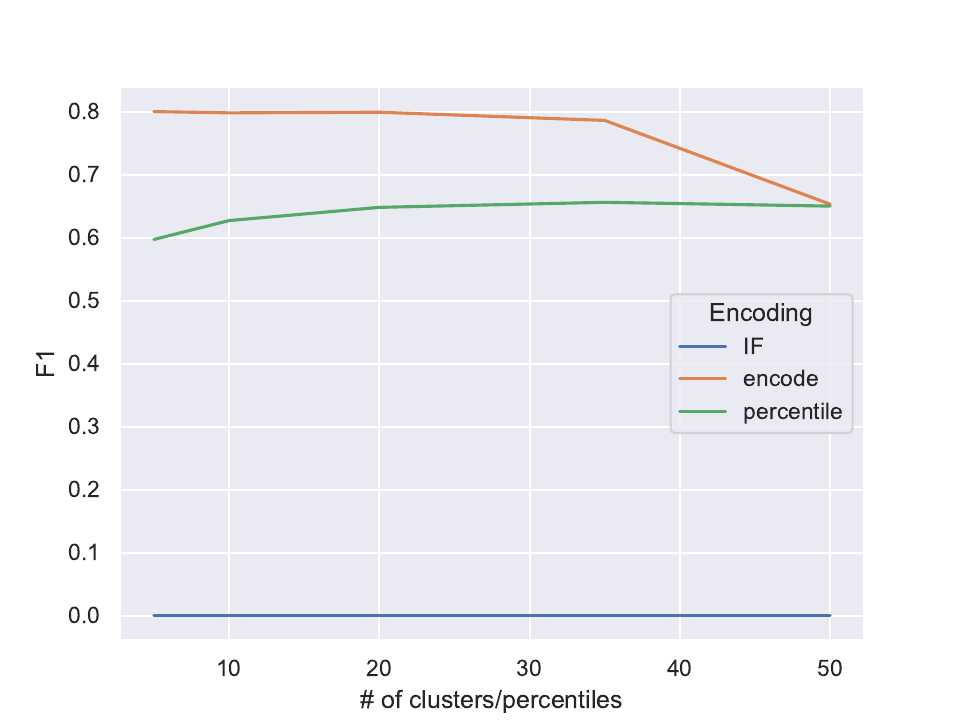}
    \includegraphics[width=0.23\linewidth]{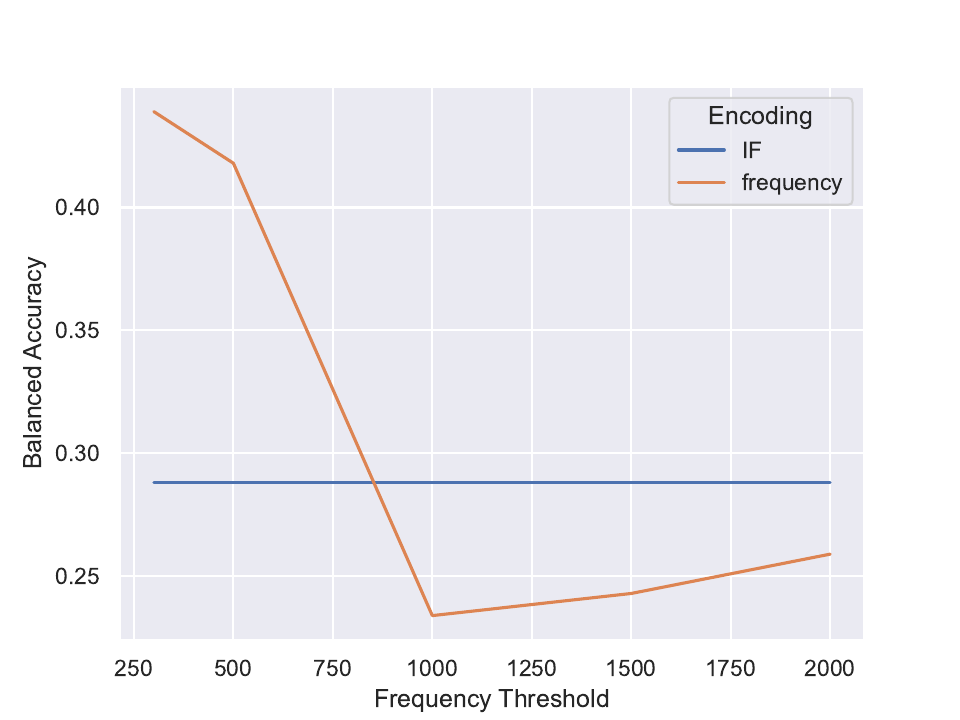}
    \includegraphics[width=0.23\linewidth]{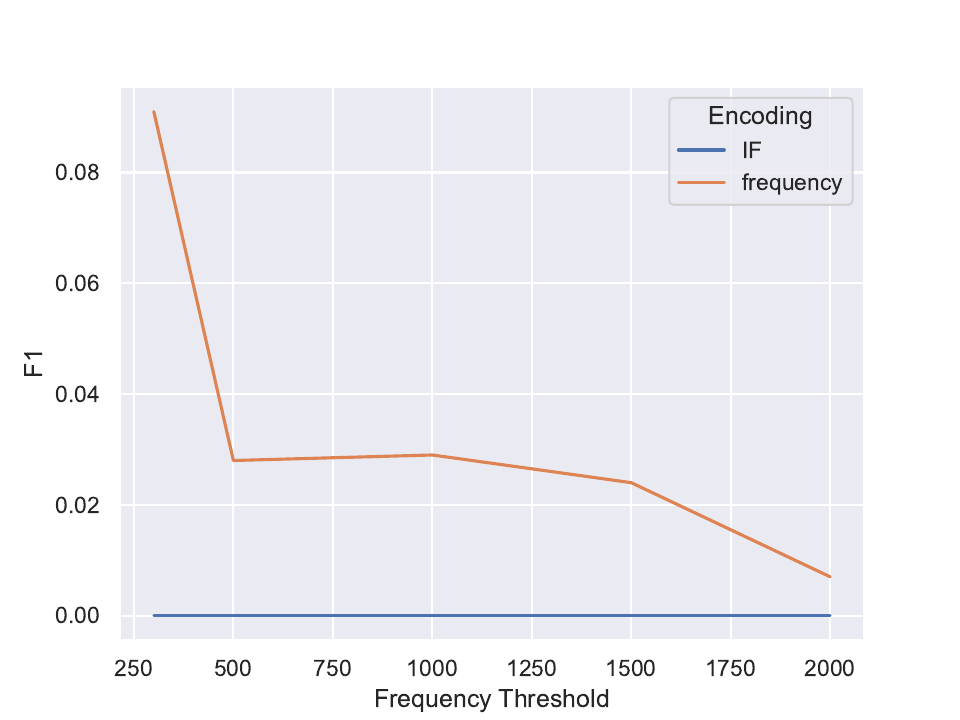}
        \caption{Performance results of IF (for each encoding method) in detecting anomalies on the AssureMOSS dataset. The figure presents the model's average results over 10 runs.}
    \label{fig:experiment_results_assuremoss_if}
\end{figure*}

\begin{figure*}[h]
    \centering
    \includegraphics[width=0.23\linewidth]{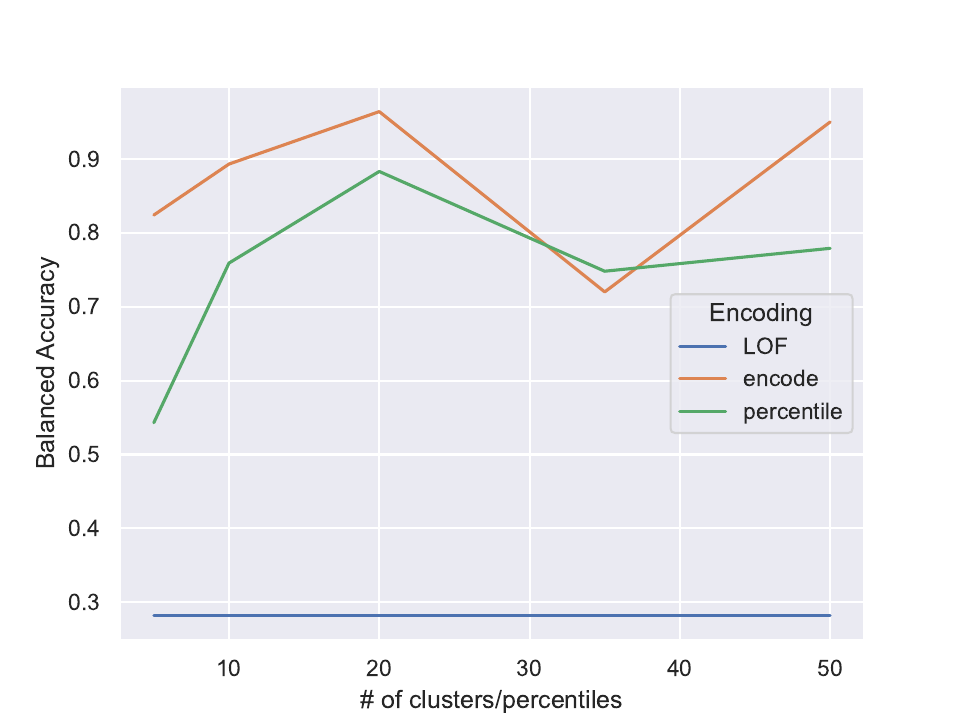}
    \includegraphics[width=0.23\linewidth]{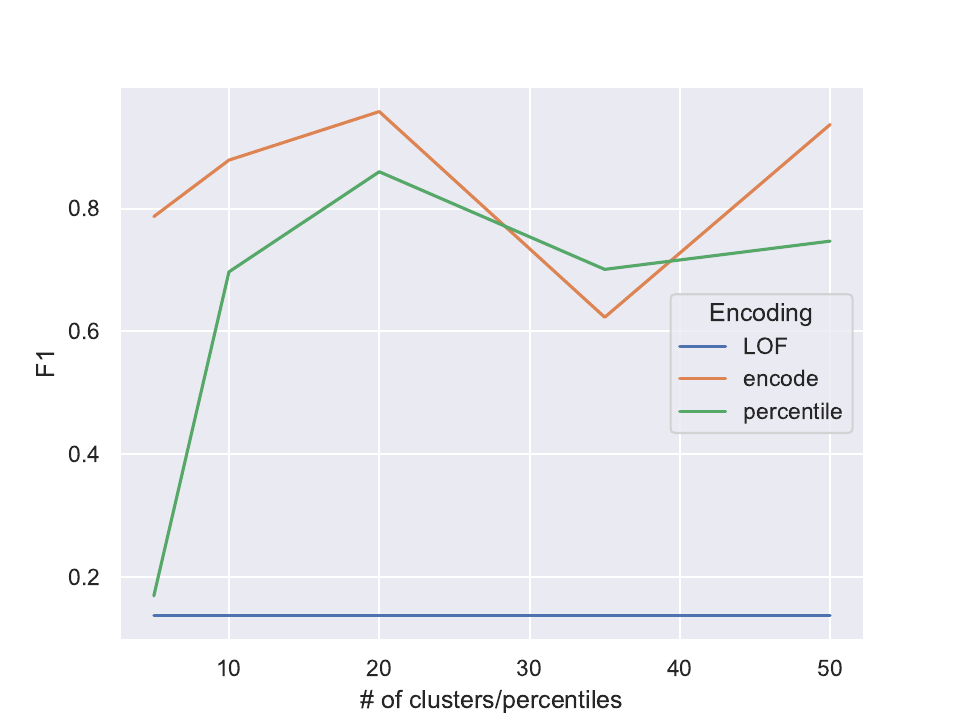}
    \includegraphics[width=0.23\linewidth]{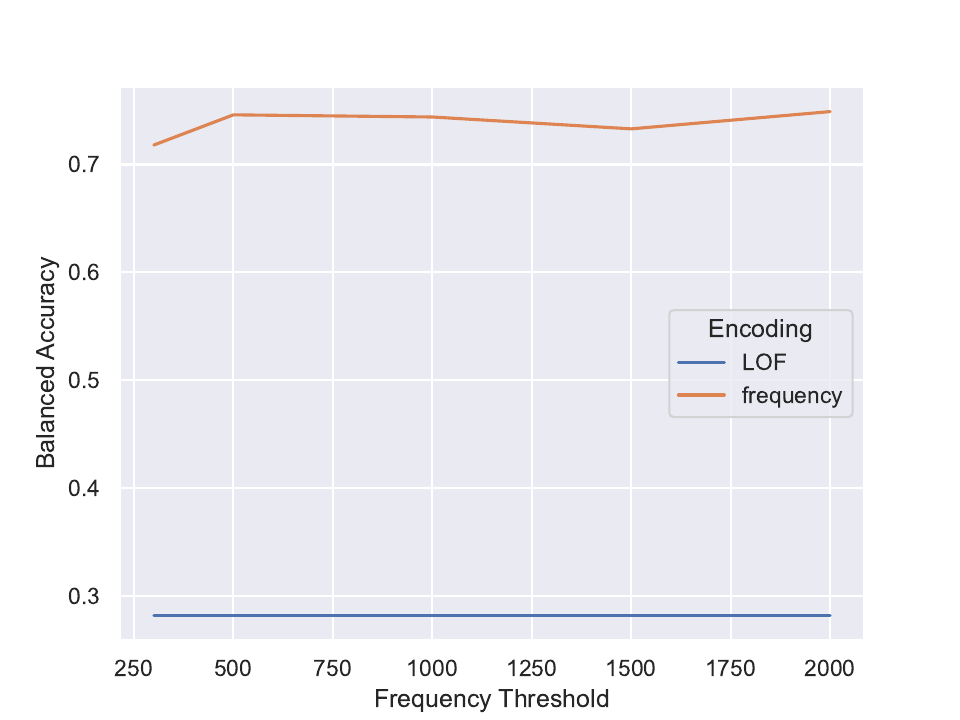}
    \includegraphics[width=0.23\linewidth]{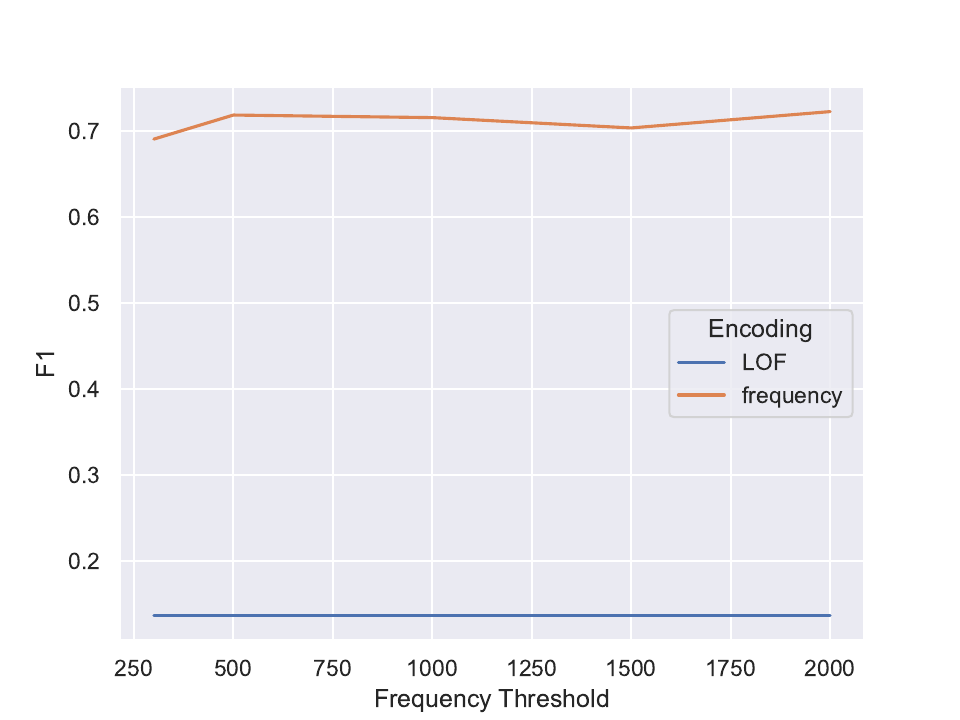}
        \caption{Performance results of LOF (for each encoding method) in detecting anomalies on the AssureMOSS dataset. The figure presents the model's average results over 10 runs.}
    \label{fig:experiment_results_assuremoss_lof}
\end{figure*}

\begin{figure*}[h]
    \centering
    \includegraphics[width=0.23\linewidth]{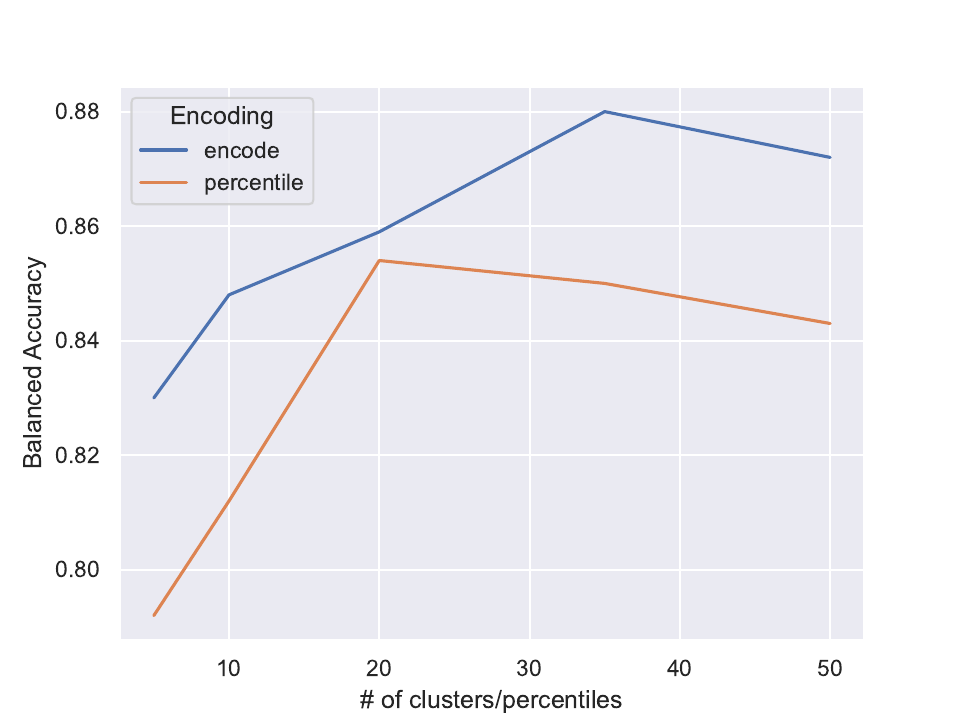}
    \includegraphics[width=0.23\linewidth]{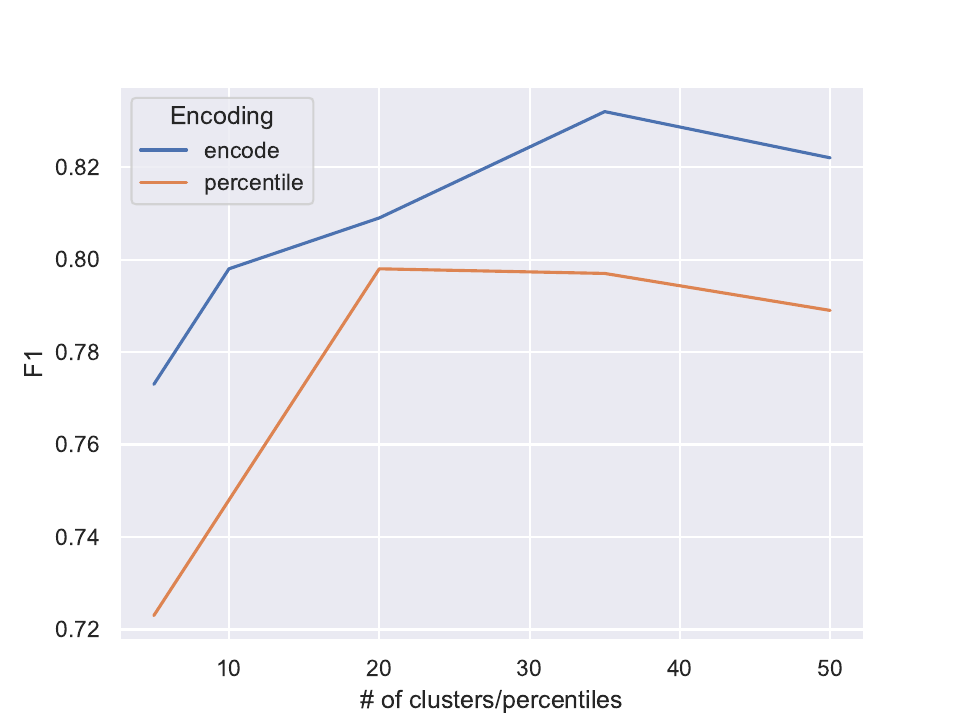}
    \includegraphics[width=0.23\linewidth]{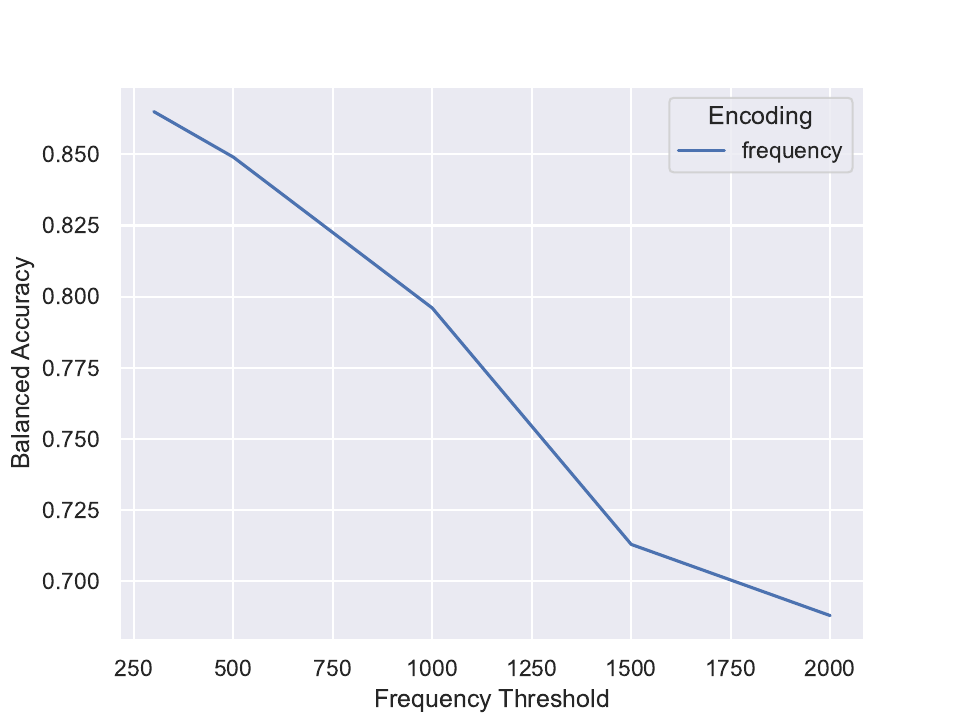}
    \includegraphics[width=0.23\linewidth]{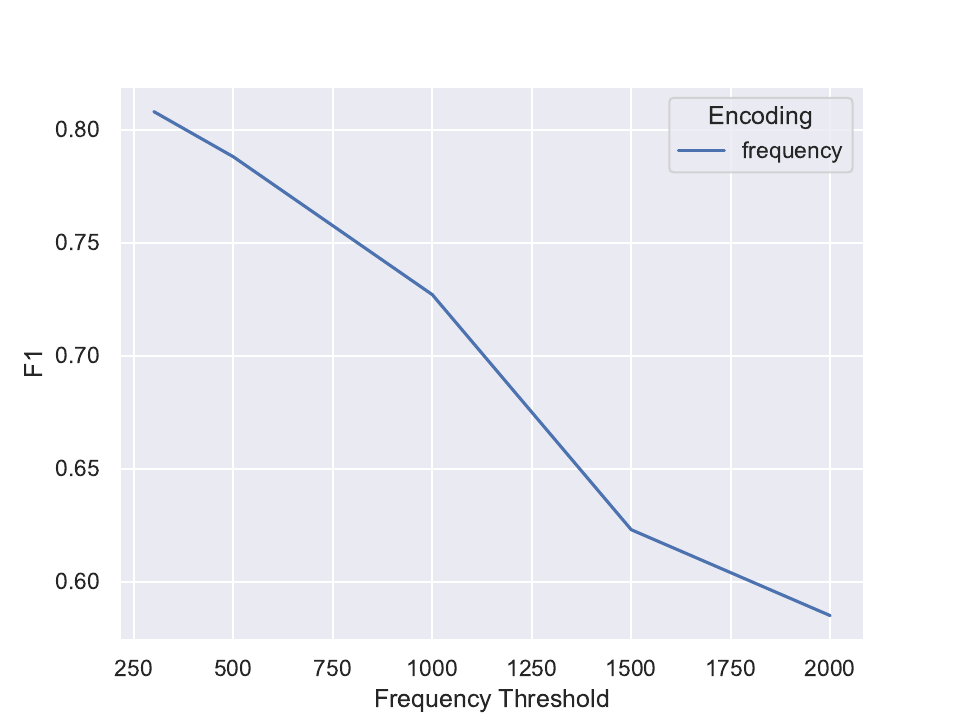}
        \caption{Performance results of State Machines (for each encoding method) in detecting anomalies on the CTU-13 dataset. The figure presents the model’s average results over 10 runs.}
    \label{fig:experiment_results_ctu_sm}
\end{figure*}

\begin{figure*}[h]
    \centering
    \includegraphics[width=0.23\linewidth]{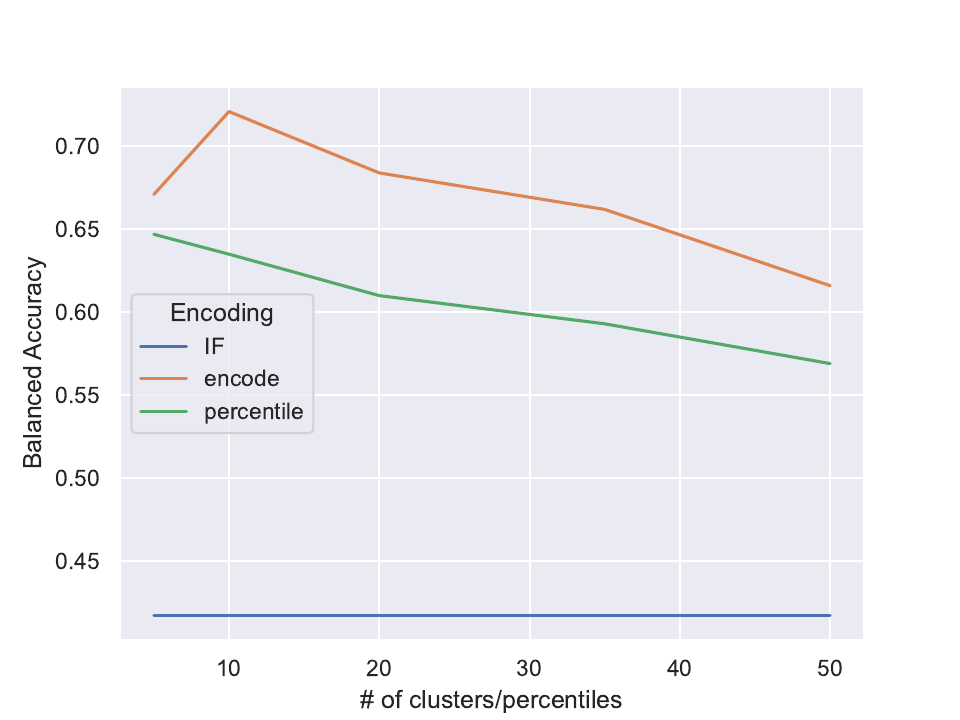}
    \includegraphics[width=0.23\linewidth]{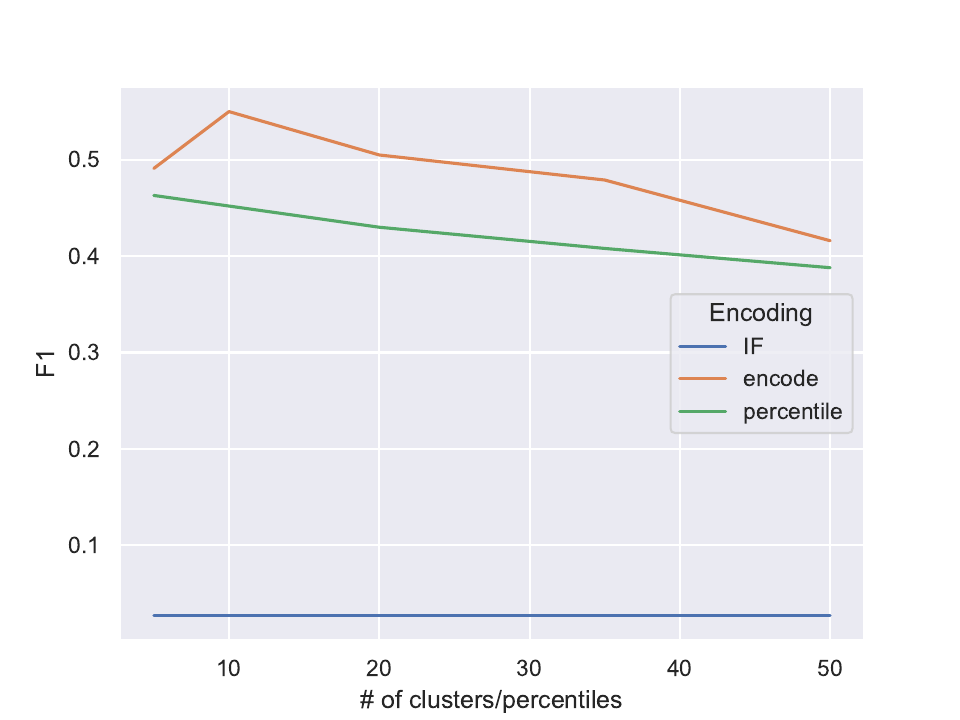}
    \includegraphics[width=0.23\linewidth]{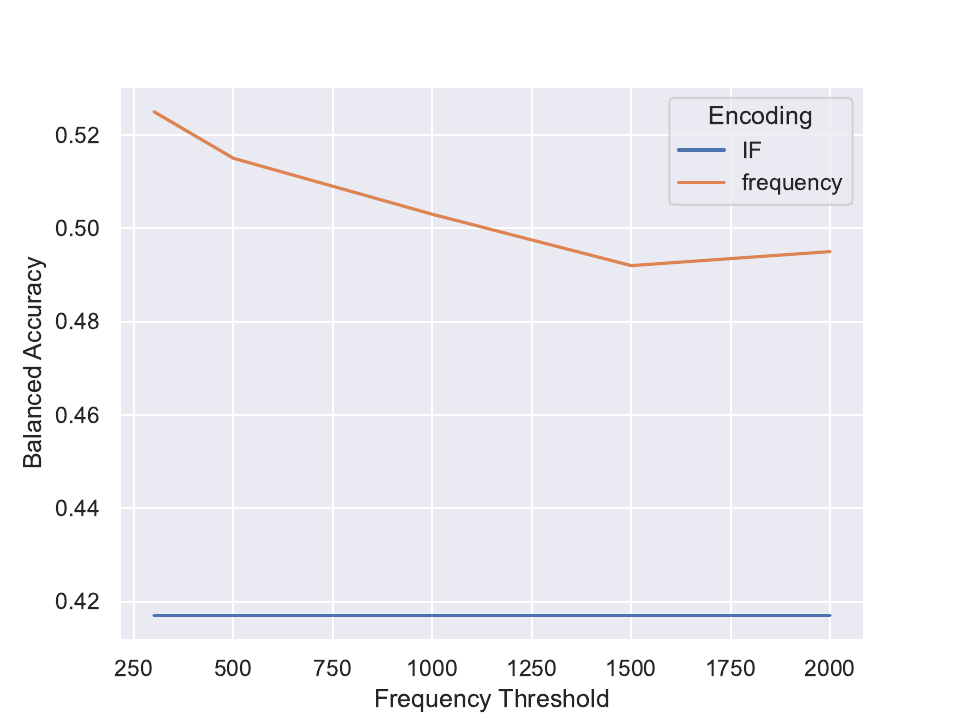}
    \includegraphics[width=0.23\linewidth]{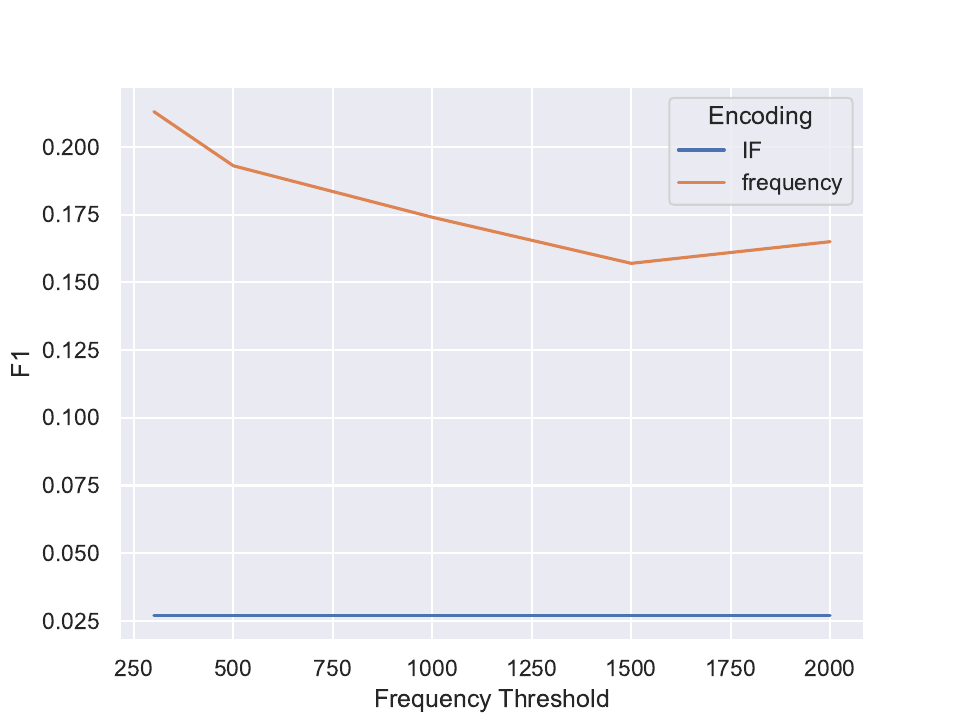}
        \caption{Performance results of IF (for each encoding method) in detecting anomalies on the CTU-13 dataset. The figure presents the model’s average results over 10 runs.}
    \label{fig:experiment_results_ctu_if}
\end{figure*}

\begin{figure*}[h]
    \centering
    \includegraphics[width=0.23\linewidth]{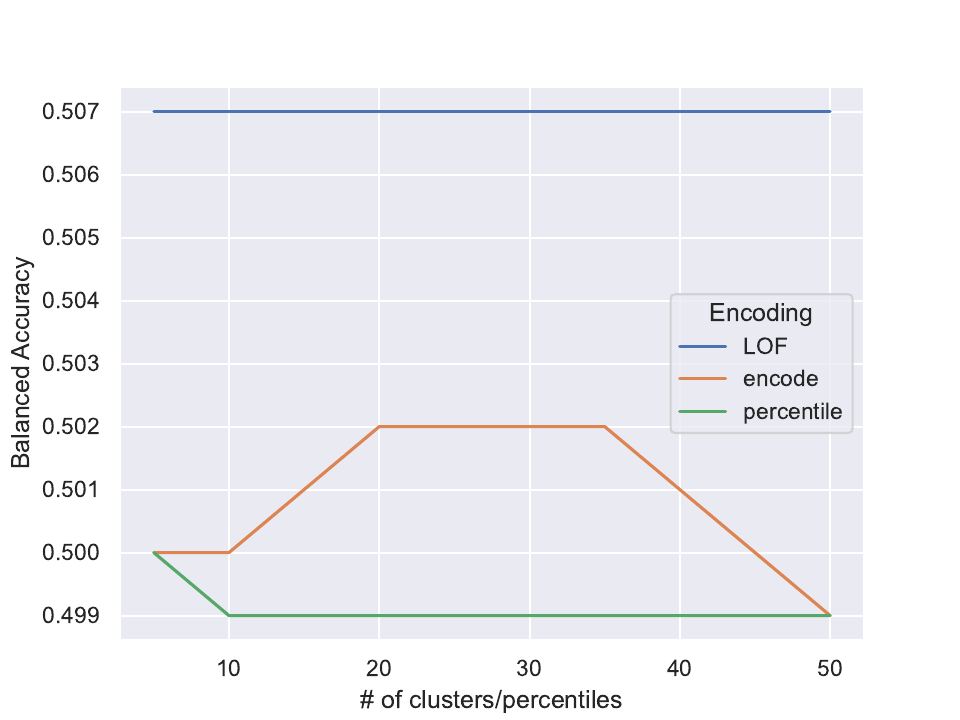}
    \includegraphics[width=0.23\linewidth]{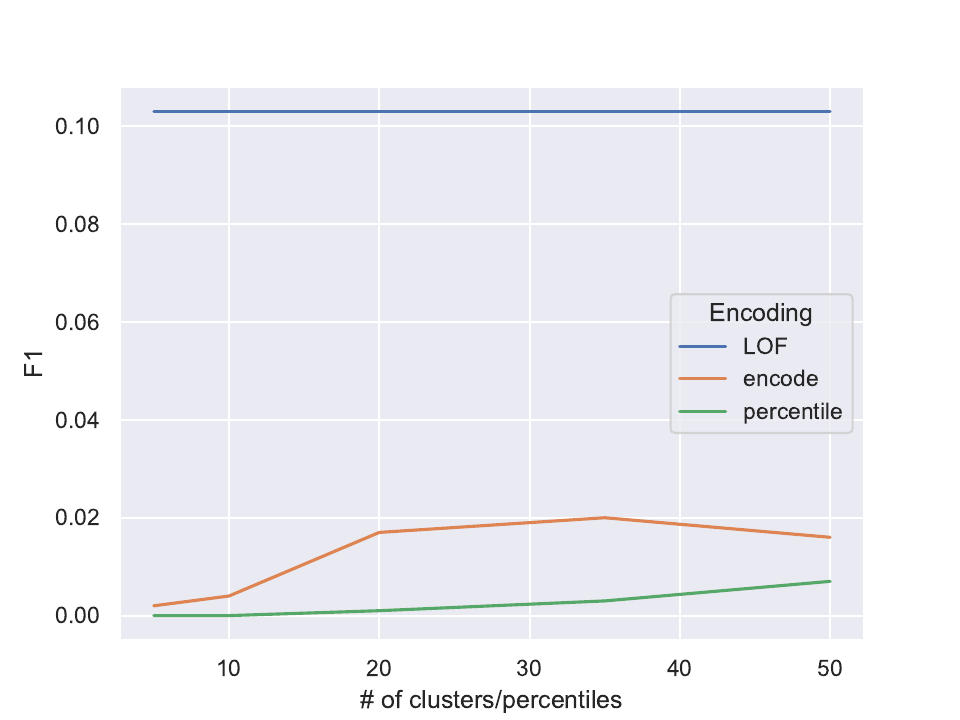}
    \includegraphics[width=0.23\linewidth]{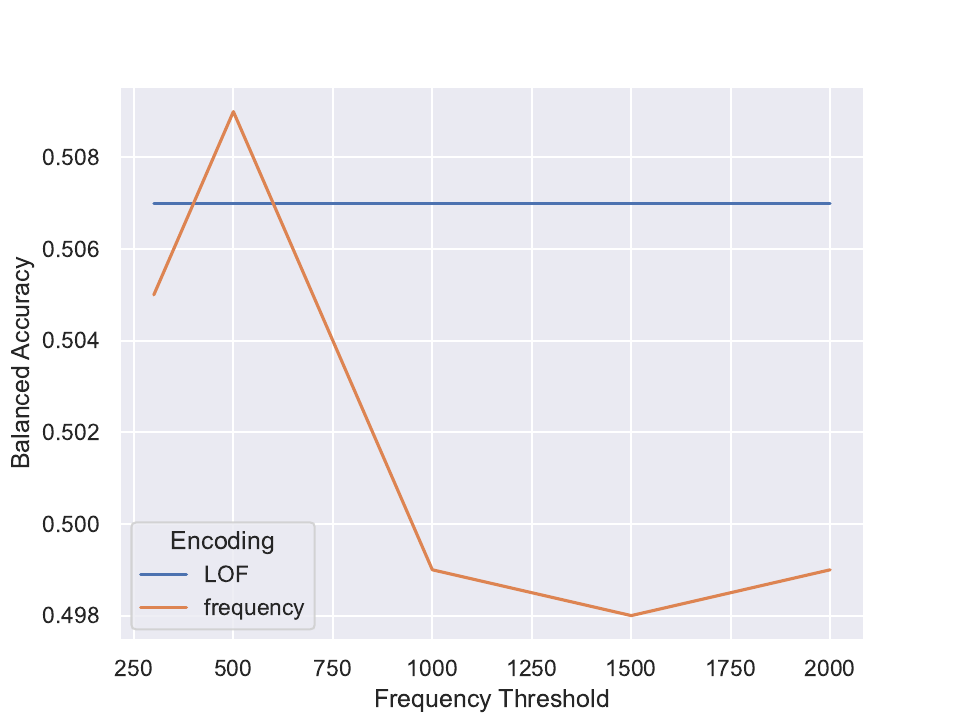}
    \includegraphics[width=0.23\linewidth]{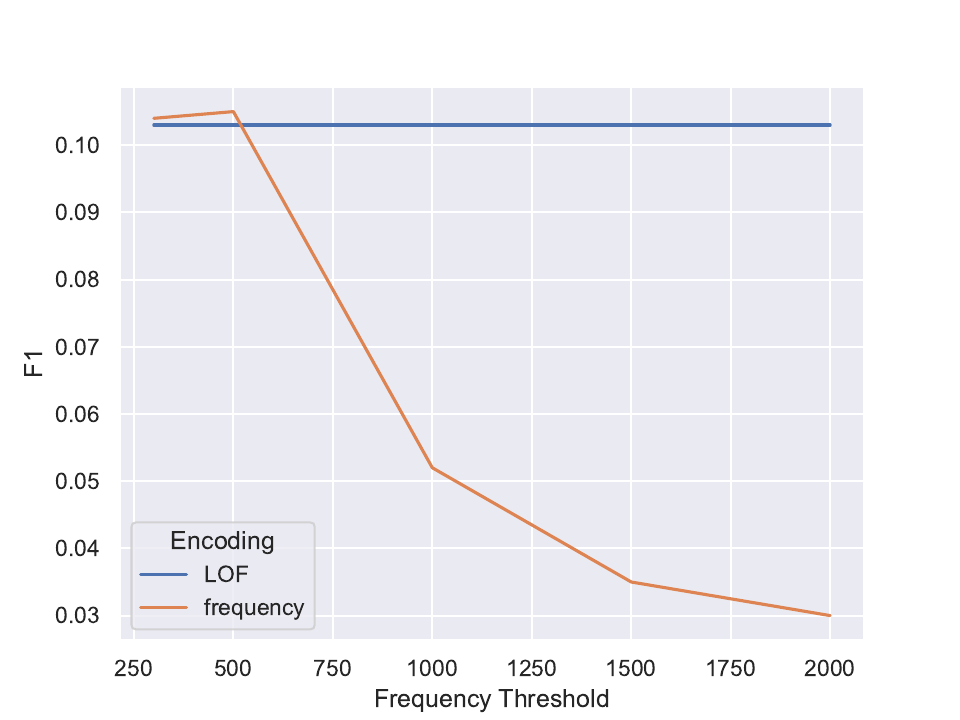}
        \caption{Performance results of LOF (for each encoding method) in detecting anomalies on the CTU-13 dataset. The figure presents the model’s average results over 10 runs.}
    \label{fig:experiment_results_ctu_lof}
\end{figure*}

\begin{figure*}[h]
    \centering
    \includegraphics[width=0.23\linewidth]{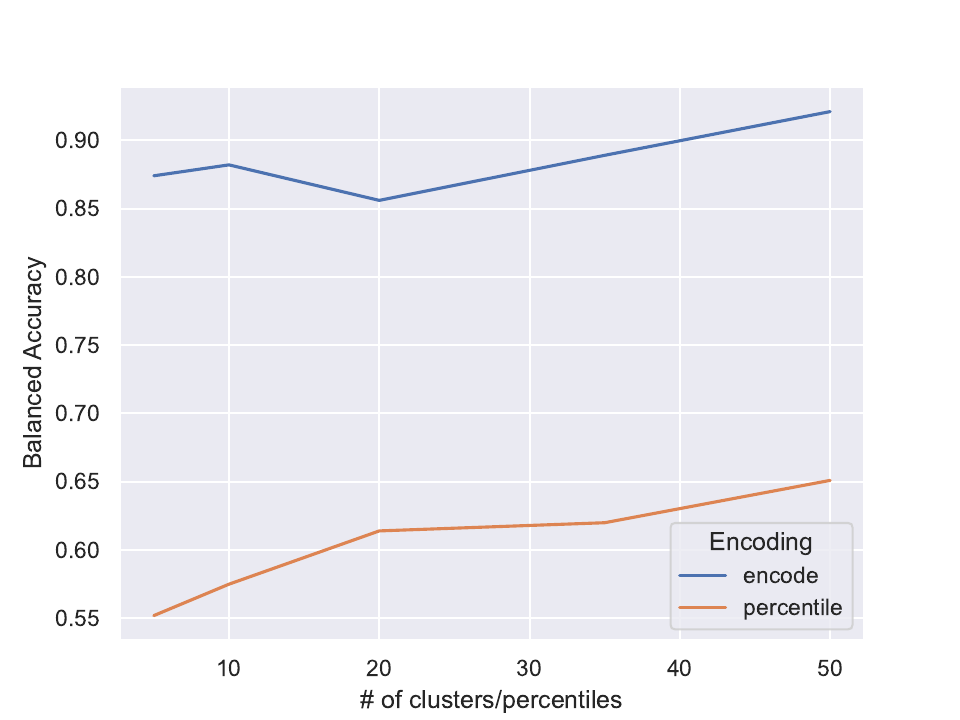}
    \includegraphics[width=0.23\linewidth]{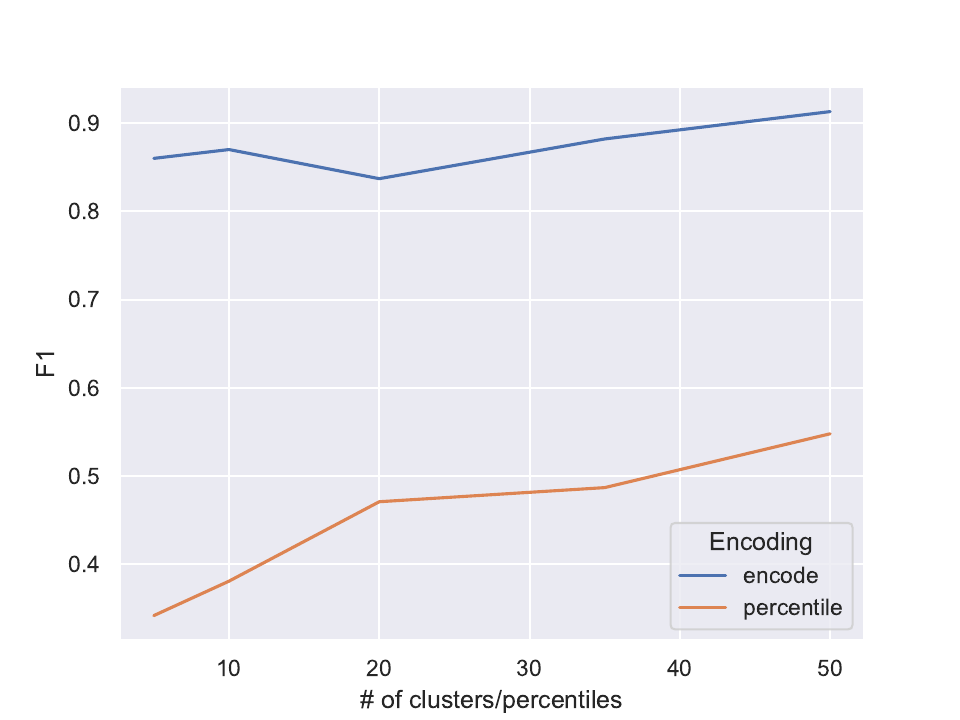}
    \includegraphics[width=0.23\linewidth]{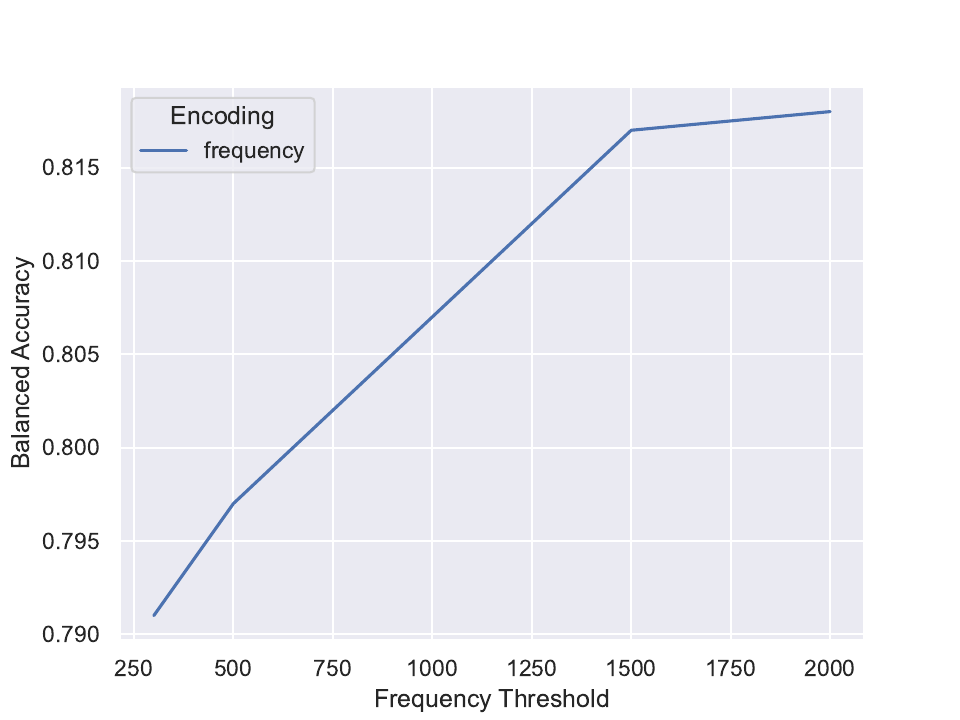}
    \includegraphics[width=0.23\linewidth]{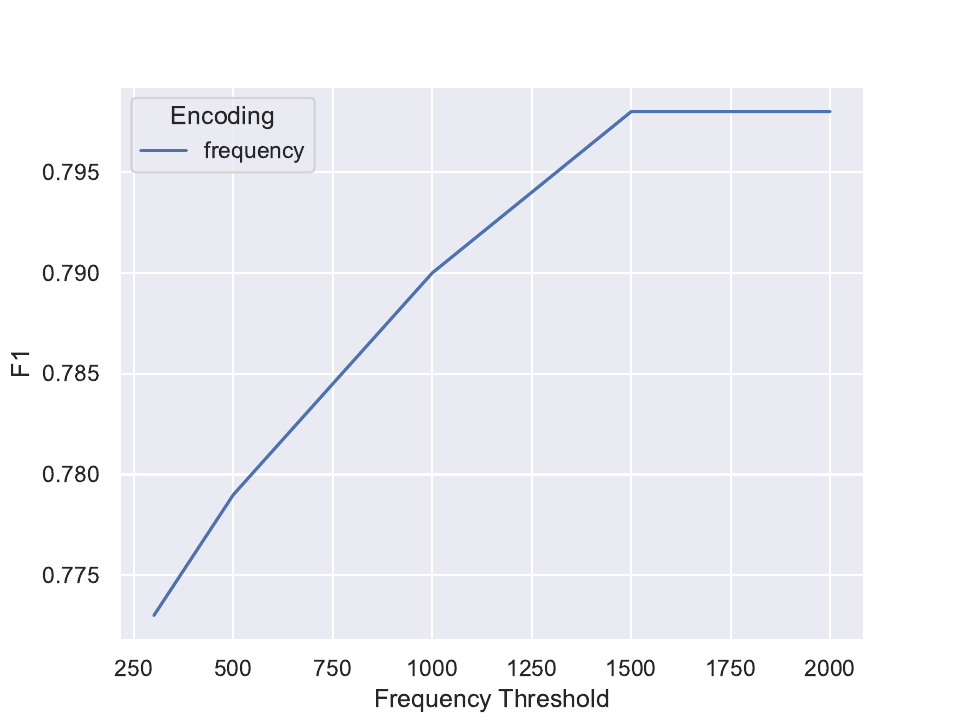}
        \caption{Performance results of State Machines (for each encoding method) in detecting anomalies on the UGR-16 dataset. The figure presents the model’s average results over 10 runs.}
    \label{fig:experiment_results_ugr_sm}
\end{figure*}

\begin{figure*}[h]
    \centering
    \includegraphics[width=0.23\linewidth]{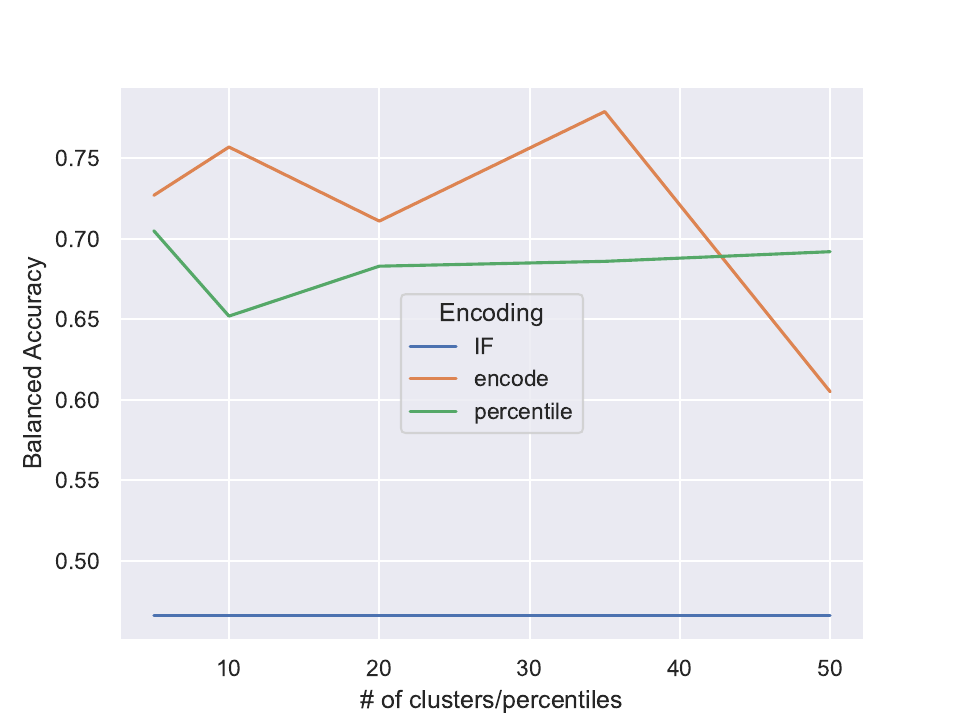}
    \includegraphics[width=0.23\linewidth]{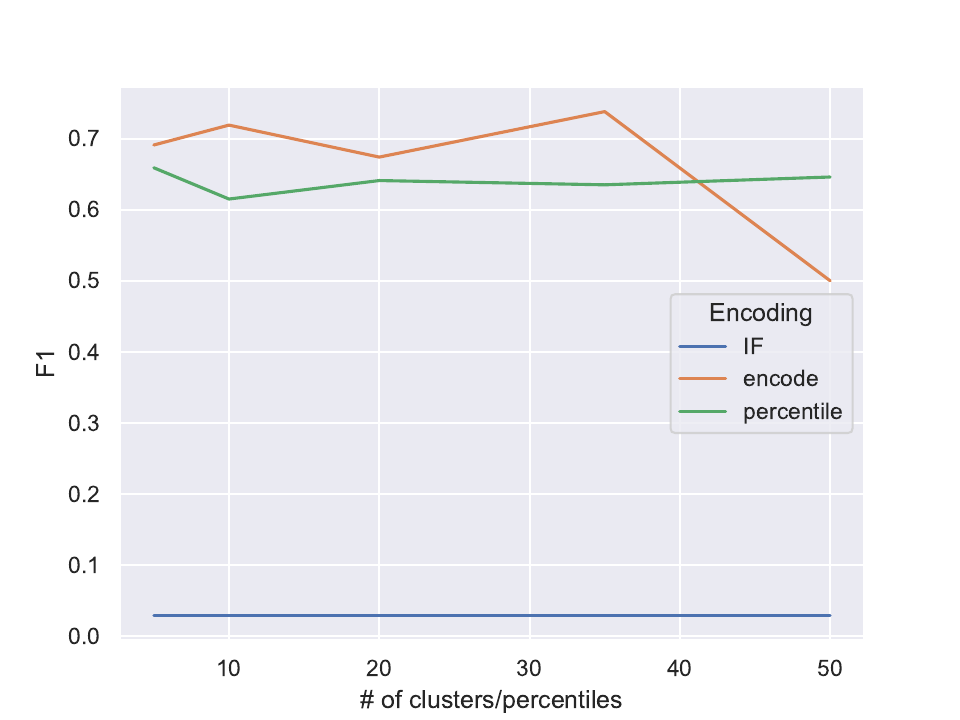}
    \includegraphics[width=0.23\linewidth]{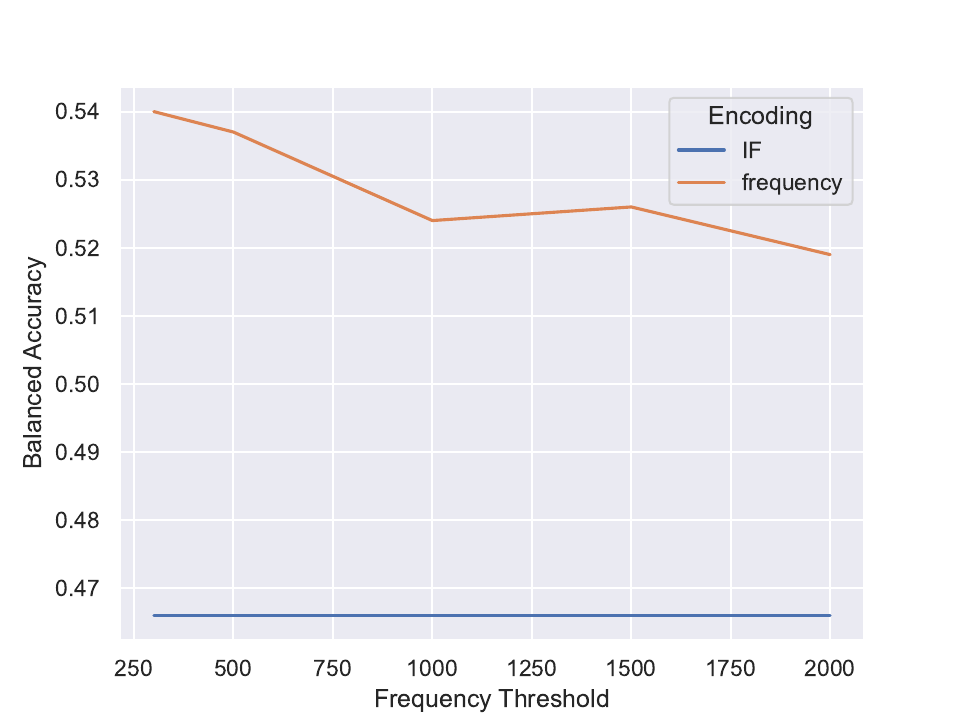}
    \includegraphics[width=0.23\linewidth]{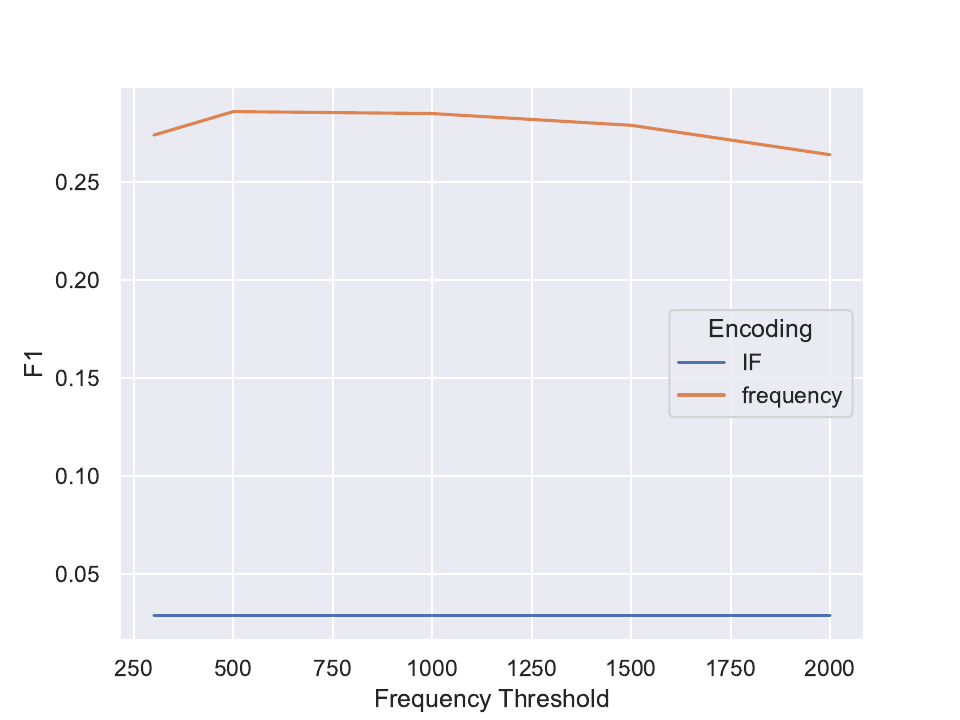}
        \caption{Performance results of IF(for each encoding method) in detecting anomalies on the UGR-16 dataset. The figure presents the model’s average results over 10 runs.}
    \label{fig:experiment_results_ugr_if}
\end{figure*}

\begin{figure*}[h]
    \centering
    \includegraphics[width=0.23\linewidth]{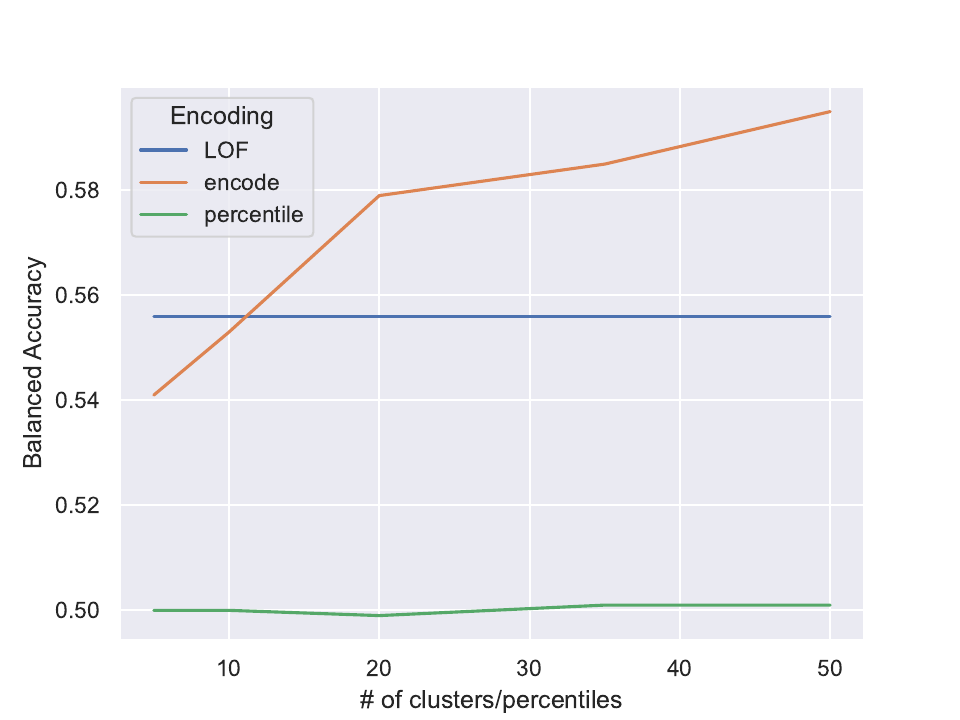}
    \includegraphics[width=0.23\linewidth]{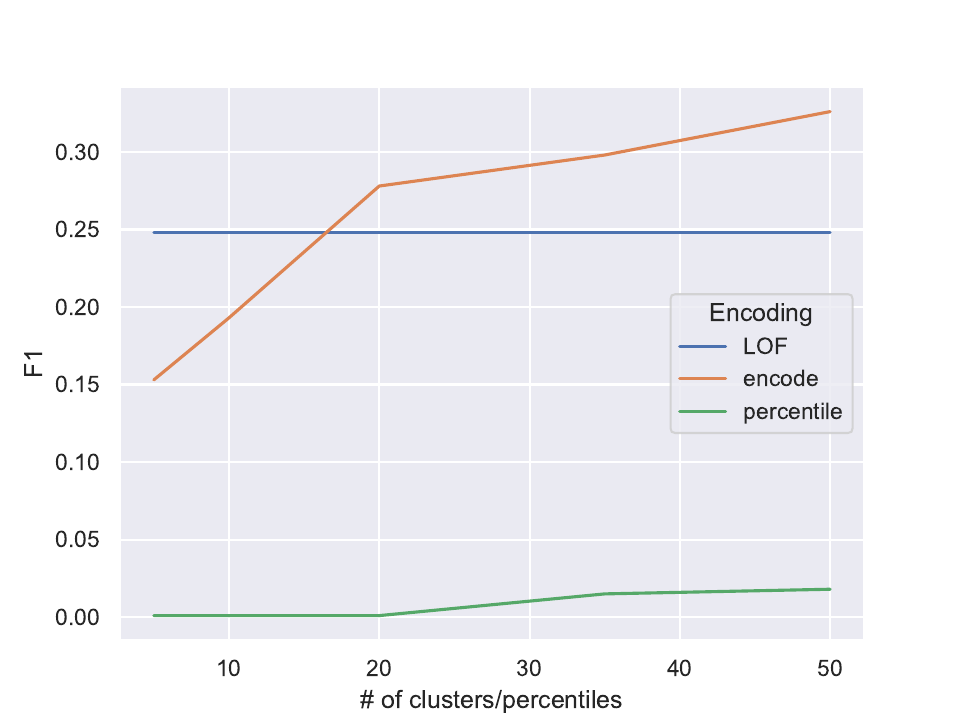}
    \includegraphics[width=0.23\linewidth]{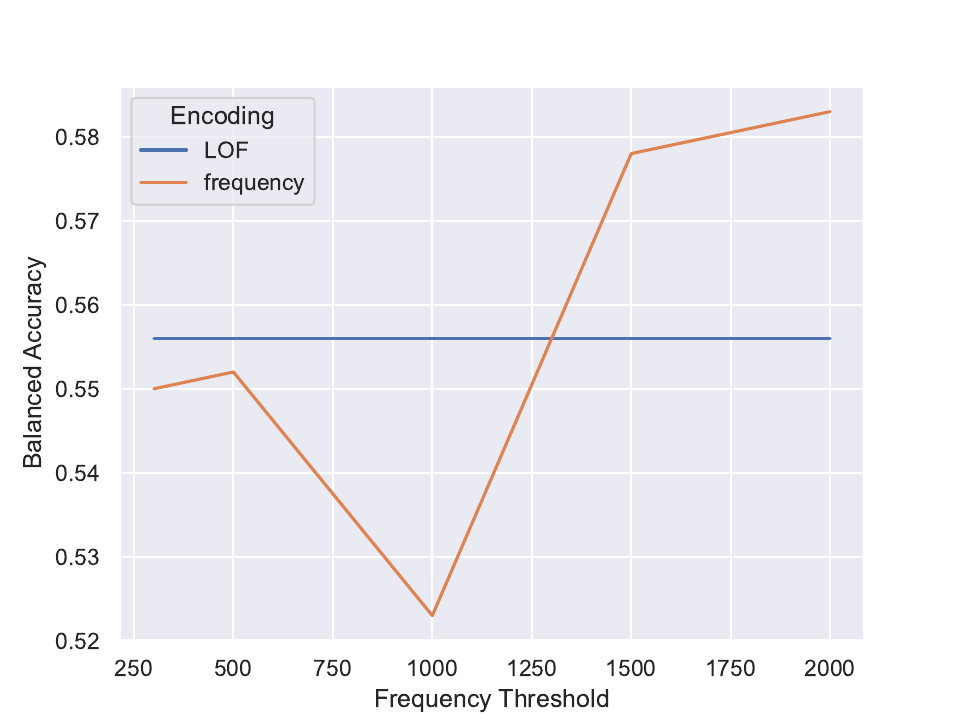}
    \includegraphics[width=0.23\linewidth]{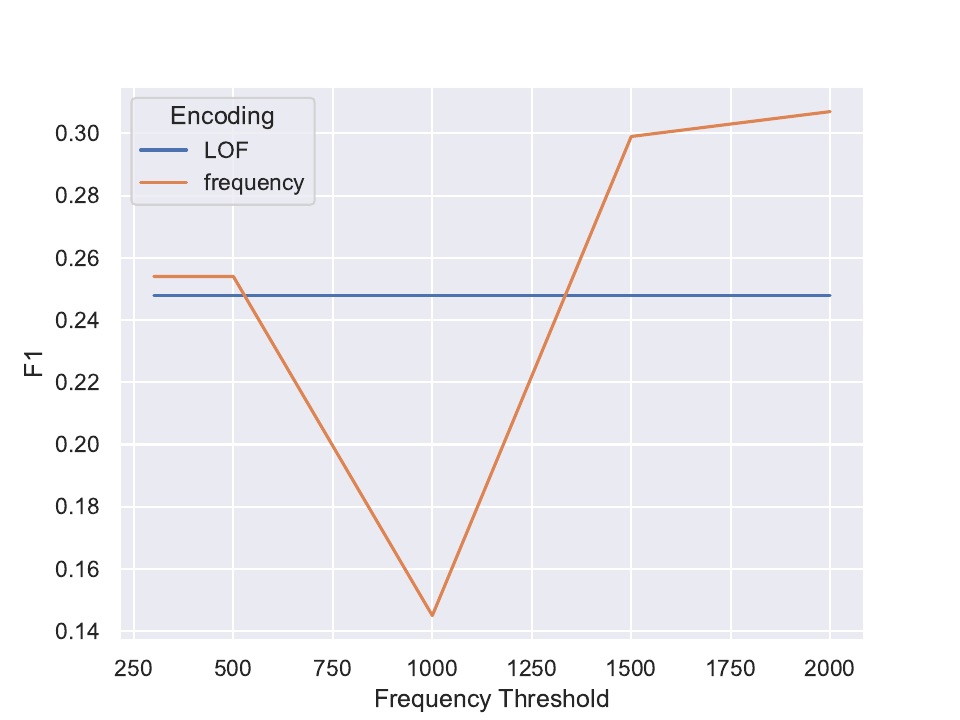}
        \caption{Performance results of LOF (for each encoding method) in detecting anomalies on the UGR-16 dataset. The figure presents the model’s average results over 10 runs.}
    \label{fig:experiment_results_ugr_lof}
\end{figure*}

In this section, we empirically demonstrate the effectiveness of ENCODE. Each selected dataset is encoded using each encoding method and following the aforementioned encoding procedure. The encoded training flows for each dataset are subsequently used to train the selected ML models for unsupervised anomaly detection. To evaluate the effectiveness of ENCODE, we compare the performance of the ML models trained on the encoded feature values provided by ENCODE with the performance of the ML models trained on encoded feature values provided by the baseline encoding methods. Figures~\ref{fig:experiment_results_assuremoss_sm}-\ref{fig:experiment_results_ugr_lof}, and Table~\ref{tab:experiment_results_assuremoss_deeplog} present the performance results of the ML models trained on feature values encoded using the three encoding methods. The performance results represent the average performance computed over ten independent runs for each ML model. The only exception is the DeepLog model, for which we were unable to compute average results due to the repeated out-of-memory issues.

\begin{table}[!ht]
\caption{Partial performance results of DeepLog on AssureMOSS dataset. Performance results are computed from a single run (due to repeated out-of-memory issues).}\label{tab:experiment_results_assuremoss_deeplog}
\begin{center}
\begin{tabular}{lccc}
\toprule
Encoding                    & Balanced Accuracy     & $F_1$\\
\midrule
ENCODE (5 clusters)         & 0.782                 & 0.734\\
Percentile (5 bins)         & 0.482                 & 0.332\\
Frequency (Threshold = 300) & -                     & -\\
\bottomrule
\end{tabular}
\end{center}
\end{table}

\subsection{Overall Results}
When we inspect the performance results presented in Figures~\ref{fig:experiment_results_assuremoss_sm}-\ref{fig:experiment_results_ugr_lof} and Table~\ref{tab:experiment_results_assuremoss_deeplog}, we observe that the models achieve considerably higher balanced accuracy and $F_1$ scores when ENCODE was used to preprocess the NetFlow data. This empirically demonstrates that the models can leverage the context defined by our encoding to learn a better contextual correlation between the feature values of the NetFlow data,  thereby improving their ability to detect anomalies within the data.  

Although the percentile-based encoding is a standard used in prior works, it performs in almost all cases worse than ENCODE. We believe this low performance is due to the fact that flow similarity is computed based on the similarity between feature values and flows with similar (feature) values do not necessarily mean that they are the same. Additionally, the percentile-based encoding does not store any contextual information on how these values co-occur.

While sequential ML models could theoretically learn such patterns, they still perform considerably better using ENCODE to preprocess NetFlow data. Since ENCODE stores contextual information within each encoded value, even non-sequential ML methods can capture some sequential information using our encoding. This effect is evident for the IF detector, which performs considerably better using our encoding than using the raw feature values. The low performance of the naive frequency-based encoding can be explained by the fact it only stores information about frequently occurring feature values (based on a given threshold), while infrequent feature values are then binned together. Similar to percentile-based encoding, this method does not capture the context of how feature values co-occur, regardless of whether they are frequent or infrequent.

When comparing the different ML methods, we observe that LOF performs worse on the CTU-13 and UGR-16 datasets when either ENCODE or percentile-based encoding is used to preprocess the NetFlow data. In fact, LOF obtained better results when it was trained using the raw feature values. LOF treats the encoded values as integers and computes distances based on them. These distances are not very meaningful, which explains the poor performance. LOF is also the only method for which it was better to the naive frequency-based encoding to preprocess the NetFlow data (see Figure~\ref{fig:experiment_results_ctu_lof}). The main reason for this observation is that the raw values remain intact for high frequencies, making the distances somewhat meaningful. Still, LOF performs considerably worse than the other two methods that can leverage our encoding in a meaningful way.

The results obtained using DeepLog are promising but are based on a single run (of one epoch) of the model on the AssureMOSS dataset. We only managed to get DeepLog working on the AssureMOSS dataset for the case where we used 5 clusters in ENCODE and 5 bins for the percentile-based encoding. We tested DeepLog both with and without GPU. In cases with a large number of clusters or bins, our machine ran out of memory. This issue is likely caused by the increase in the number of unique symbols introduced when the size of the clusters or bins increases (more clusters or bins means that we have more symbols). Although the results show that DeepLog achieves considerably better performance using ENCODE to preprocess the NetFlow data, future work is needed to determine the significance of this improvement.

\subsection{The Effect of Cluster Number On Model Performance}
A key parameter of ENCODE is the number of clusters used in KMeans clustering. The larger this number, the more distinctions are made between flows, but at a cost of increased complexity. 
Based on the performance results presented in Figures~\ref{fig:experiment_results_assuremoss_sm}-\ref{fig:experiment_results_ugr_lof}, we observe that increasing the number of clusters can varying effects on the performance of the ML models: (1) it can increase the model's performance (see Figure~\ref{fig:experiment_results_assuremoss_lof}, Figure~\ref{fig:experiment_results_ctu_sm} and Figure~\ref{fig:experiment_results_ugr_sm}), (2) it can decrease the model's performance (see Figure~\ref{fig:experiment_results_assuremoss_if}, Figure~\ref{fig:experiment_results_ctu_if} and Figure\ref{fig:experiment_results_ugr_if}), or (3) have no considerable effect on the model's performance (see~\ref{fig:experiment_results_assuremoss_sm}. The increase in performance appears to be more common with state machines, while the decrease in performance is more frequent in IFs. This highlights the importance of tuning this parameter within ENCODE to determine the optimal number of clusters for a given model to achieve a performance boost. 

\section{Use-case and robustness}\label{sec:robustness_encode}
\begin{figure}[!ht]
  \centering
  \subfloat[Likelihood values computed for training data of UGR-16 dataset.]{
    \includegraphics[width=0.45\textwidth]{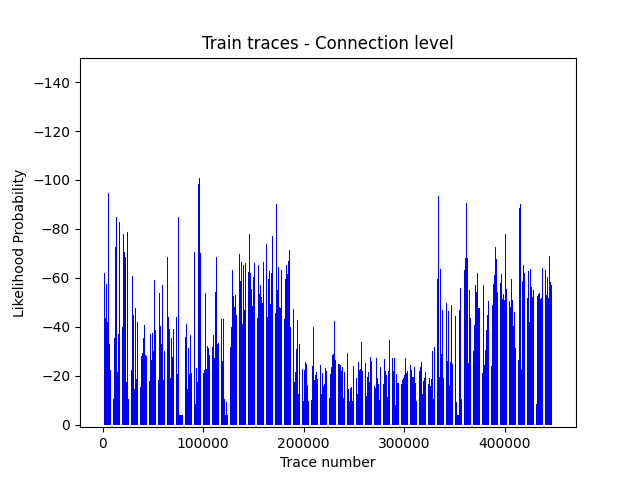}
  }
  \\
  \subfloat[Likelihood values computed for test data of UGR-16 dataset.]{
    \includegraphics[width=0.45\textwidth]{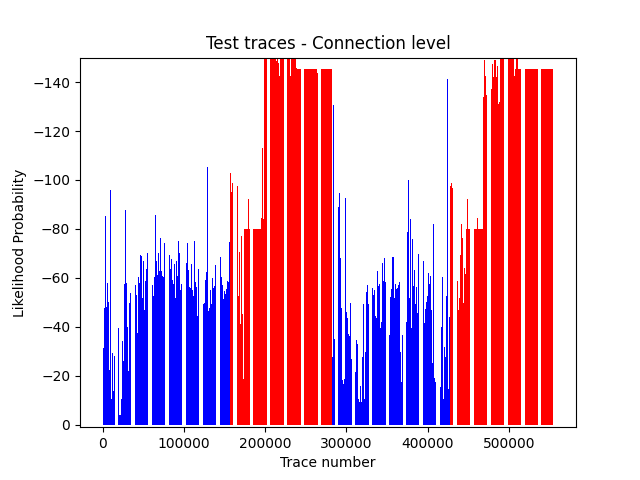}
  }
  \caption{Likelihood values computed by the state machine model for the UGR-16 dataset. Blue vertical lines represent benign traces, while red vertical lines represent malicious traces.}\label{fig:ctu_loglikelihood}
\end{figure}

From our experimental results, we observe using 35 clusters provides the best performance for state machines. As ENCODE was originally developed for state machine learning, we now present a use case demonstrating how ENCODE can be used to derive deeper insights into the model’s predictions. To illustrate this, we display the output an analyst would see when they are running an anomaly detection system that utilizes state machines. The state machine produces (log)likelihood values, which the anomaly detection system can use to determine the degree of abnormality for an arbitrary trace. In our experiments, we use a fixed decision threshold, tuned on the training data, to count false/true positives and negatives. However, in real-world applications, we believe it is more effective to visualize the likelihood values over time rather than relying solely on threshold-based decisions. Figure~\ref{fig:ctu_loglikelihood} presents example visualizations of likelihood values generated for the UGR-16 dataset.

The likelihood values of the traces are visualized over time, enabling an analyst to quickly determine whether an alert corresponds to an attack or a false alarm. We often observe a large cluster of low-likelihood values (represented by tall vertical lines) when malicious traces occur. Additionally, we observe some anomalous traces in the train data simply because these connections are infrequent. The likelihood values for the test data tend to be lower (indicating a higher anomaly score) when new, infrequent flows occur. The figure also demonstrates that, although the performance values in our experiments are computed on a per-flow basis, it is more practical to visualize these values over time to observe (considerable) changes in network behavior. One could even argue that the method only triggers a limited number of false alarms, as indicated by the small number of blue peaks in the test data. Furthermore, it is evident that a single attack can produce multiple peaks (tall verticle lines) in the likelihood scores, even though just one peak would be sufficient to trigger an investigation

\subsection{Experimental Results}
To test the robustness of ENCODE combined with state machines for anomaly detection, we present the performance of the state machine model after introducing various levels of perturbation to the number of bytes and packets in the test data (see Table~\ref{tab:experiment_results_ctu_perturbed}). Three types of perturbations were applied in our experiments: (1) adding a fixed value to both features, (2) adding a random value between 0\% and 50\% of the original feature values, and (3) adding a random value between 50\% and 80\% of the original feature values.Applying these perturbations to the feature values causes them to be assigned to the cluster that groups all infrequent values (with low co-occurrence). Since these perturbed feature values are grouped together with other infrequent values, it becomes difficult to derive meaningful distances and contextual relationships from these values. Interestingly, the performance of the state machine model improves when perturbations are added to the features. This suggests that our encoding, when combined with a state machine model, is robust against attackers who attempt to randomly or structurally modify their network traffic to evade detection.

\begin{table}[!ht]
	\caption{Performance results of state machines on perturbed test data. }\label{tab:experiment_results_ctu_perturbed}
	\begin{center}
        \resizebox{\columnwidth}{!}{
		\begin{tabular}{lccccc}
			\toprule
			Amount of \\Perturbation                              & Balanced Accuracy      & $F_1$     & Precision    & Recall\\
			\midrule
			Fixed 10               & 0.915                 & 0.860      & 0.754        & 1.0\\
			Random between 0\% and 50\%   & 0.915                 & 0.860      & 0.754        & 1.0\\
			Random between 50\% and 80\%   & 0.915                 & 0.860      & 0.754        & 1.0\\
			No perturbations    & 0.850                 & 0.797      & 0.754         & 0.844 \\
			\bottomrule
		\end{tabular}
            }
		\label{tab1}
	\end{center}
\end{table}

\section{Conclusion \& Future Work}\label{sec:conclusion_future_encode}
We present a new encoding algorithm, ENCODE, that preprocesses NetFlow datasets for machine learning. ENCODE uses the frequency of feature values and the co-occurrence frequencies between feature values to compute the context in which the feature values appear. This context is then used to group feature values, creating an encoding for the given input feature. The intuition is that feature values that occur in similar contexts and with similar frequencies are indicative of similar software services and should, therefore, receive the same encoding label.

We empirically demonstrate the effectiveness of ENCODE by applying it to NetFlow data from three different datasets and training four different ML models to detect network anomalies in an unsupervised manner. For each dataset, we evaluated the performance results of the ML models and compared them to the performance achieved using two baseline encoding methods. The empirical results show that the ML models performed considerably better when trained on NetFlows preprocessed by ENCODE. 

From our empirical results, the state machine model achieved the best performance in detecting network anomalies. For this method, we present a use case that demonstrates the type of information an analyst receives from the system and how it can be used to perform further analysis. Additionally, we show that this method is resilient to perturbations introduced by attackers attempting to evade detection. In fact, such perturbations make the data appear more anomalous, resulting in higher detection rates.

In this work, we have applied ENCODE to NetFlow data. In future work, we aim to evaluate ENCODE on other types of data, such as software log data. We believe that our encoding could also be beneficial for preprocessing software log data, and it would be valuable to investigate its effectiveness on additional data types.

Furthermore, it would be interesting to evaluate ENCODE on a broader range of ML models. For instance, although our primary focus is unsupervised learning, we believe our encoding could also be useful in supervised learning tasks.

Finally, we are developing a streaming version of ENCODE that can be deployed in real-time with NetFlow collectors.

\section*{Acknowledgments}
This work is funded under the Assurance and certification in secure
Multi-party Open Software and Services (AssureMOSS) Project,
(https://assuremoss.eu/en/), with the support of the European Commission and H2020 Program, under Grant Agreement No. 952647.

\balance
\bibliographystyle{plain}
\bibliography{references}

\begin{thebibliography}{10}

\bibitem{assureMOSS_project}
{AssureMOSS}.
\newblock \url{https://assuremoss.eu/en/}.

\bibitem{deeplog_implementation_repo}
{DeepLog - PyTorch implementation of Deeplog: Anomaly detection and diagnosis
  from system logs through deep learning}.
\newblock \url{https://github.com/Thijsvanede/DeepLog}.

\bibitem{encode_repo}
{ENCODE: Encoding NetFlows for Network Anomaly Detection}.
\newblock \url{https://github.com/tudelft-cda-lab/ENCODE}.

\bibitem{microsoft_xp_ports}
{Service overview and network port requirements - Windows Server | Microsoft
  Docs}.
\newblock
  \url{https://docs.microsoft.com/en-us/troubleshoot/windows-server/networking/service-overview-and-network-port-requirements}.

\bibitem{sklearn_kmeans}
{sklearn.cluster.KMeans — scikit-learn 1.1.1 documentation}.
\newblock
  \url{https://scikitlearn.org/stable/modules/generated/sklearn.cluster.KMeans.html}.

\bibitem{sklearn-if}
{sklearn.ensemble.IsolationForest — scikit-learn 1.1.2 documentation}.
\newblock
  \url{https://scikit-learn.org/stable/modules/generated/sklearn.ensemble.\\IsolationForest.html}.

\bibitem{sklearn-silhouette}
{sklearn.metrics.silhouette\_score — scikit-learn 1.1.2 documentation}.
\newblock
  \url{https://scikit-learn.org/stable/modules/generated/sklearn.metrics.silhouette\_score.html}.

\bibitem{sklearn-lof}
{sklearn.neighbors.LocalOutlierFactor — scikit-learn 1.1.2 documentation}.
\newblock
  \url{https://scikitlearn.org/stable/modules/generated/sklearn.\\neighborsLocalOutlierFactor.html}.

\bibitem{Ahmed2020_deep_learning}
Abdulghani~Ali Ahmed, Waheb~A. Jabbar, Ali~Safaa Sadiq, and Hiran Patel.
\newblock {Deep learning-based classification model for botnet attack
  detection}.
\newblock {\em Journal of Ambient Intelligence and Humanized Computing},
  1:1--10, mar 2020.

\bibitem{Camacho2019_semi_supervised}
Jose Camacho, Gabriel Macia-Fernandez, Noemi~Marta Fuentes-Garcia, and Edoardo
  Saccenti.
\newblock {Semi-supervised multivariate statistical network monitoring for
  learning security threats}.
\newblock {\em IEEE Transactions on Information Forensics and Security},
  14(8):2179--2189, aug 2019.

\bibitem{assuremoss_dataset}
Clinton Cao and Agathe Blaise.
\newblock {AssureMOSS Kubernetes Run-time Monitoring Dataset}.
\newblock \url{https://data.4tu.nl/articles/\_/20463687/1}, 2022.

\bibitem{Cao_2022_Learning}
Clinton Cao, Agathe Blaise, Sicco Verwer, and Filippo Rebecchi.
\newblock Learning state machines to monitor and detect anomalies on a
  kubernetes cluster.
\newblock In {\em Proceedings of the 17th International Conference on
  Availability, Reliability and Security}, ARES '22, New York, NY, USA, 2022.
  Association for Computing Machinery.

\bibitem{CiscoSystems2018}
{Cisco Systems}.
\newblock {Cisco IOS NetFlow - Cisco}.
\newblock
  \url{https://www.cisco.com/c/en/us/products/ios-nx-os-software/ios-netflow/index.html},
  2018.

\bibitem{du2017deeplog}
Min Du, Feifei Li, Guineng Zheng, and Vivek Srikumar.
\newblock Deeplog: Anomaly detection and diagnosis from system logs through
  deep learning.
\newblock In {\em Proceedings of the 2017 ACM SIGSAC conference on computer and
  communications security}, pages 1285--1298, 2017.

\bibitem{Garcia2014_an_empirical}
S.~Garcia, M.~Grill, J.~Stiborek, and A.~Zunino.
\newblock {An empirical comparison of botnet detection methods}.
\newblock {\em Computers \& Security}, 45:100--123, sep 2014.

\bibitem{Grov_2019_Towards}
Gudmund Grov, Wei Chen, Marc Sabate, and David Aspinall.
\newblock Towards intelligible robust anomaly detection by learning
  interpretable behavioural models.
\newblock In {\em Vol. 12 (2019): NISK 2019; Proceedings of the 12th Norwegian
  Information Security Conference}, NISK. Akademika, November 2019.
\newblock 12th Norwegian Information Security Conference : Co-located with
  NIKT, NISK 2019 ; Conference date: 25-11-2019 Through 27-11-2019.

\bibitem{Haghighat2021_intrusion_detection}
Mohammad~Hashem Haghighat and Jun Li.
\newblock {Intrusion detection system using voting-based neural network}.
\newblock {\em Tsinghua Science and Technology}, 26(4):484--495, aug 2021.

\bibitem{Larriva-Novo2020_efficient}
Xavier Larriva-Novo, Mario Vega-Barbas, V{\'{i}}ctor~A. Villagr{\'{a}}, Diego
  Rivera, Manuel {\'{A}}lvarez-Campana, and Julio Berrocal.
\newblock {Efficient Distributed Preprocessing Model for Machine Learning-Based
  Anomaly Detection over Large-Scale Cybersecurity Datasets}.
\newblock {\em Applied Sciences 2020, Vol. 10, Page 3430}, 10(10):3430, may
  2020.

\bibitem{Liu08_isolation}
Fei~Tony Liu, Kai~Ming Ting, and Zhi-Hua Zhou.
\newblock Isolation forest.
\newblock In {\em 2008 Eighth IEEE International Conference on Data Mining},
  pages 413--422, 2008.

\bibitem{Macia-Fernandez2018_ugr}
Gabriel Maci{\'{a}}-Fern{\'{a}}ndez, Jos{\'{e}} Camacho, Roberto
  Mag{\'{a}}n-Carri{\'{o}}n, Pedro Garc{\'{i}}a-Teodoro, and Roberto
  Ther{\'{o}}n.
\newblock {UGR‘16: A new dataset for the evaluation of
  cyclostationarity-based network IDSs}.
\newblock {\em Computers and Security}, 73:411--424, mar 2018.

\bibitem{Magan-Carrion2020towards}
Roberto Mag{\'{a}}n-Carri{\'{o}}n, Daniel Urda, Ignacio D{\'{i}}az-Cano, and
  Bernab{\'{e}} Dorronsoro.
\newblock {Towards a Reliable Comparison and Evaluation of Network Intrusion
  Detection Systems Based on Machine Learning Approaches}.
\newblock {\em Applied Sciences 2020, Vol. 10, Page 1775}, 10(5):1775, mar
  2020.

\bibitem{Matousek_2021_Efficient}
Petr Matoušek, Vojtěch Havlena, and Lukáš Holík.
\newblock Efficient modelling of ics communication for anomaly detection using
  probabilistic automata.
\newblock In {\em 2021 IFIP/IEEE International Symposium on Integrated Network
  Management (IM)}, pages 81--89, 2021.

\bibitem{microsoft_threat_matrix}
Microsoft.
\newblock {Threat matrix for Kubernetes}.
\newblock
  \url{https://www.microsoft.com/security/blog/2020/04/02/attack-matrix-kubernetes/},
  2020.

\bibitem{Mikolov2013_efficient}
Tomas Mikolov, Kai Chen, Greg Corrado, and Jeffrey Dean.
\newblock {Efficient estimation of word representations in vector space}.
\newblock In {\em 1st International Conference on Learning Representations,
  ICLR 2013 - Workshop Track Proceedings}. International Conference on Learning
  Representations, ICLR, jan 2013.

\bibitem{Moustafa2015_unsw}
Nour Moustafa and Jill Slay.
\newblock {UNSW-NB15: A comprehensive data set for network intrusion detection
  systems (UNSW-NB15 network data set)}.
\newblock {\em 2015 Military Communications and Information Systems Conference,
  MilCIS 2015 - Proceedings}, dec 2015.

\bibitem{Nguyen2019_gee}
Quoc~Phong Nguyen, Kar~Wai Lim, Dinil~Mon Divakaran, Kian~Hsiang Low, and
  Mun~Choon Chan.
\newblock {GEE: A Gradient-based Explainable Variational Autoencoder for
  Network Anomaly Detection}.
\newblock {\em 2019 IEEE Conference on Communications and Network Security, CNS
  2019}, pages 91--99, jun 2019.

\bibitem{Nugraha2020_performance}
Beny Nugraha, Anshitha Nambiar, and Thomas Bauschert.
\newblock {Performance Evaluation of Botnet Detection using Deep Learning
  Techniques}.
\newblock {\em Proceedings of the 11th International Conference on Network of
  the Future, NoF 2020}, pages 141--149, oct 2020.

\bibitem{cic-ids}
University of~New~Brunswick.
\newblock {Datasets | Research | Canadian Institute for Cybersecurity | UNB}.
\newblock \url{https://www.unb.ca/cic/datasets/index.html}.

\bibitem{Pellegrino2017_learning}
Gaetano Pellegrino, Qin Lin, Christian Hammerschmidt, and Sicco Verwer.
\newblock {Learning behavioral fingerprints from Netflows using Timed
  Automata}.
\newblock In {\em Proceedings of the IM 2017 - 2017 IFIP/IEEE International
  Symposium on Integrated Network and Service Management}, pages 308--316.
  Institute of Electrical and Electronics Engineers Inc., jul 2017.

\bibitem{Piskozub2019_malalert}
Michal Piskozub, Riccardo Spolaor, and Ivan Martinovic.
\newblock {MalAlert: Detecting malware in large-scale network traffic using
  statistical features}.
\newblock In {\em Performance Evaluation Review}, volume~46, pages 151--154,
  2019.

\bibitem{Shamshirband2019_a_new_malware}
Shahab Shamshirband and Anthony~T. Chronopoulos.
\newblock {A new malware detection system using a high performance-elm method}.
\newblock {\em ACM International Conference Proceeding Series}, jun 2019.

\bibitem{Terzi2017_big_data}
Duygu~Sinanc Terzi, Ramazan Terzi, and Seref Sagiroglu.
\newblock {Big data analytics for network anomaly detection from netflow data}.
\newblock In {\em 2nd International Conference on Computer Science and
  Engineering, UBMK 2017}, pages 592--597. Institute of Electrical and
  Electronics Engineers Inc., oct 2017.

\bibitem{Verwer2017_Flexfringe}
Sicco Verwer and Christian~A. Hammerschmidt.
\newblock {Flexfringe: A passive automaton learning package}.
\newblock In {\em Proceedings - 2017 IEEE International Conference on Software
  Maintenance and Evolution, ICSME 2017}, pages 638--642, 2017.

\bibitem{Yilmaz2019_expansion}
Ibrahim Yilmaz and Rahat Masum.
\newblock {Expansion of Cyber Attack Data From Unbalanced Datasets Using
  Generative Techniques}.
\newblock 2019.

\bibitem{Zoppi2021_unsupervised}
Tommaso Zoppi, Andrea Ceccarelli, Tommaso Capecchi, and Andrea Bondavalli.
\newblock Unsupervised anomaly detectors to detect intrusions in the current
  threat landscape.
\newblock {\em ACM/IMS Trans. Data Sci.}, 2(2), apr 2021.

\bibitem{Zoppi2021_meta_learning}
Tommaso Zoppi, Mohamad Gharib, Muhammad Atif, and Andrea Bondavalli.
\newblock {Meta-learning to improve unsupervised intrusion detection in
  cyber-physical systems}.
\newblock {\em ACM Transactions on Cyber-Physical Systems}, 5(4), sep 2021.

\end{thebibliography}






\end{document}